\documentclass[12pt]{article}

\usepackage{macros}

\usepackage[margin = .95in]{geometry}

\pdfoutput=1

\usepackage{amsmath,amssymb,color}
\usepackage{url}

\usepackage{algorithmic}
\usepackage{algorithm2e}

\usepackage{subfigure}

\usepackage{xcolor}

\def\X{\mathbf{X}}

\def\e{\mathbf{e}}
\def\R{\mathbb{R}}

\def\I{\mathbf{I}}
\def\x{\mathbf{x}}
\def\v{\mathbf{v}}
\def\u{\mathbf{u}}

\def\u{\mathbf{u}}

\def\mmu{\boldsymbol{\mu}}
\def\sgn{\mathrm{sgn}}
\def\hmmu{\hat{\boldsymbol{\mu}}}

\title{\bf Clustering Large Data Sets with Incremental Estimation of Low-density Separating Hyperplanes}

\begin{document}

\author{\name David P.\ Hofmeyr \hfill
{\small \textmd{ Department of Statistics and Actuarial Science}}\\
\textcolor{white}{.}\hfill {\small \textmd{Stellenbosch University}}\\
\textcolor{white}{.}\hfill {\small \textmd{7600, South Africa}}
}

\maketitle

\begin{abstract}%
An efficient method for obtaining low-density hyperplane separators in the unsupervised context is proposed. Low density separators can be used to obtain a partition of a set of data based on their allocations to the different sides of the separators. The proposed method is based on applying stochastic gradient descent to the integrated density on the hyperplane with respect to a convolution of the underlying distribution and a smoothing kernel. In the case where the bandwidth of the smoothing kernel is decreased towards zero, the bias of these updates with respect to the true underlying density tends to zero, and convergence to a minimiser of the density on the hyperplane can be obtained. A post-processing of the partition induced by a collection of low-density hyperplanes yields an efficient and accurate clustering method which is capable of automatically selecting an appropriate number of clusters.
%A by-product of the determination of low density hyperplanes is a set of projections which can be used for the purpose of dimension reduction, and the proposed approach offers an alternative to existing incremental dimension reduction methods in the unsupervised domain.
Experiments with the proposed approach show that it is highly competitive in terms of both speed and accuracy when compared with relevant benchmarks. Code to implement the proposed approach is available in the form of an {\tt R} package from \url{https://github.com/DavidHofmeyr/iMDH}.
\end{abstract}


%% -- Introduction -------------------------------------------------------------

%% - In principle "as usual".
%% - But should typically have some discussion of both _software_ and _methods_.
%% - Use {}, \pkg{}, and \code{} markup throughout the manuscript.
%% - If such markup is in (sub)section titles, a plain text version has to be
%%   added as well.
%% - All software mentioned should be properly \cite-d.
%% - All abbreviations should be introduced.
%% - Unless the expansions of abbreviations are proper names (like "Journal
%%   of Statistical Software" above) they should be in sentence case (like
%%   "generalized linear models" below).

\section{Introduction}
\label{sec:intro}

The guiding principle in the non-parametric statistical approach to clustering, known as density based clustering, is that clusters form connected regions of high density in the underlying probability distribution and are separated by regions of relatively low density. While most popular approaches to density clustering focus on these high density regions directly~\citep{ester1996density, rinaldo2010generalized}, the Minimum Density Hyperplane (MDH) approach focuses instead on the low-density regions which separate them~\citep{pavlidis2016JMLR, hofmeyrPPCI}.  In this work we will investigate the estimation of minimum density hyperplanes in the fully incremental setting where access to only a single datum at each iteration is required. This allows for the application of the MDH framework to very large data sets, in terms of both number and dimensionality.

A hyperplane in $\R^d$ is a translated subspace of co-dimension one and may be described by all points, $\x\in\R^d$, which satisfy the linear equation $\v^\top \x = b$, where $\v \in \R^d, ||\v|| = 1$, and $b\in \R$ parameterise the hyperplane, i.e.,
\begin{align}
    H(\v, b) := \{\x \in \R^d | \v^\top \x = b\}.
\end{align}
The density on a hyperplane, $H(\v, b)$, with respect to a continuous probability distribution on $\R^d$ with density function $f(\cdot)$ is given by the surface integral
\begin{align}\label{eq:denhyp}
    I_f(\v, b) := \oint_{\x: \v^\top\x = b} f(\x) \ d\x.
\end{align}
Of great practical convenience is that this integral is simply equal to the density of the random variable $\v^\top X$ evaluated at $b$, where $X$ has density $f(\cdot)$. That is, if we introduce the general notation $f_{Z}(\cdot)$ to represent the density function of an arbitrary continuous random variable $Z$, then $I_{f_X}(\v, b) =  f_{\v^\top X}(b)$. As a consequence the above integral can be estimated efficiently using, for example, a kernel estimate from a sample of realisations of $X$ projected onto the vector $\v$~\citep{pavlidis2016JMLR}.

A hyperplane in $\R^d$ forms a binary partition of any set of points in $\R^d$, say $\{\x_1, ..., \x_n\}$, based on the sides of the hyperplane on which each of the points lies. That is, separating those $\x_i$'s for which $\v^\top\x_i \geq 0$ from those for which $\v^\top\x_i < 0$. Multiple hyperplanes can be combined to produce a more refined clustering either in a hierarchical structure~\citep{boley1998principal, tasoulis2010enhancing, hofmeyr2017PAMI} or by taking the intersections of multiple binary partitions formed by each of the hyperplanes~\citep{pena2001cluster}. Although numerous methods have been proposed for obtaining high quality hyperplanes for clustering in the offline/batch context~\citep{pena2001cluster, hofmeyr2015maximum, pavlidis2016JMLR, hofmeyr2017PAMI, wang2020efficient}, the problem has received very little attention in the online/incremental setting. Low density hyperplanes orthogonal to incrementally estimated principal components have been used with reasonable success~\citep{tasoulis2012clustering, hofmeyr2016divisive}, however the limitations of principal components for this task have been well documented~\citep{pena2001cluster, pavlidis2016JMLR, wang2020efficient}. In this paper we explore the problem of estimating minimum density hyperplanes in the fully incremental setting. In particular, we apply a modified stochastic gradient descent~(SGD) to minimise the integral given in~(\ref{eq:denhyp}) by taking the convolution of a sequence of i.i.d. random variables with a smoothing kernel. Since the bias of a kernel density estimator is independent of the sample size, close-to unbiased estimates of the gradient of~(\ref{eq:denhyp}) can be obtained using only a single observation at a time.
%
%If the bandwidth in the smoothing kernel converges to some non-zero lower bound then this SGD converges almost surely to a minimiser of~(\ref{eq:denhyp}) where $f(\cdot)$ is given by the convolution of the streaming density and the smoothing kernel. On the other hand, i
%
If the bandwidth decreases to zero at an appropriate rate, convergence to a minimiser of~(\ref{eq:denhyp}) where $f(\cdot)$ is the true underlying density can be achieved.

In order to obtain a complete clustering model, we adopt a hierarchical framework in which a complete bisecting tree model of chosen depth is estimated. The binary partitions at each internal node in the tree are determined by a minimum density hyperplane estimated from the subset of observations allocated to the node. 
%
%The entire model can be estimated incrementally. 
%
To obtain a final solution the model is then pruned using the common within-cluster sum of squares objective and the subtree selected is determined using an elbow technique. This approach yields highly accurate models for clustering of large and potentially high dimensional data sets. A further advantage of hierarchical clustering models of this sort is that they lend themselves to better interpretability and can be subjectively validated using low dimensional projections of the observations allocated to each of the nodes which expose the cluster separation therein~\citep{hofmeyrPPCI}.

\iffalse

A useful by-product of finding a set of minimum density hyperplanes, say $H(\v_1, b_1), ..., H(\v_k, b_k)$, is that the collection of vectors $\v_1, ..., \v_k$ can be used to perform dimension reduction by taking the projections of a set of observations in $\R^d$ onto these vectors. Since minimum density hyperplanes are designed to pass between high density clusters, if these observations arise from the same distribution as that from which the minimum density hyperplanes were estimated, then the projections of these observations will tend to reveal much of their cluster/modal structure.

\fi

The remainder of this paper is organised as follows. In the following section we discuss the proposed approach for incrementally estimating minimum density hyperplanes. We give an overview of the convergence analysis, but leave some technical details to the appendix. In Section~\ref{sec:clustering} we describe the construction of a hierarchical clustering model in greater detail, and discuss explicitly the pruning step. In Section~\ref{sec:experiments} we present results from a set of experiments designed to investigate the performance of the proposed approach, in comparison with existing incremental and offline clustering methods for large data sets. %We also provide a subjective analysis of the performance of the method for the purpose of dimension reduction, by way of low-dimensional visualisations of the projected data.

\section{Incremental Estimation of Minimum Density Hyperplanes}\label{sec:convergence}

In this section we discuss a simple update scheme which can be used to estimate minimum density hyperplanes in a fully incremental setting. We assume that we receive a sequence, $X^{(1)}, X^{(2)}, ...$, of independent random variables identically distributed to $X$, with continuous distribution function $F_X(\cdot)$ and corresponding density function $f_X(\cdot)$.
%
%As is common~\citep{CCIPCA} we assume that $\mmu := E[X^{(1)}] = \mathbf{0}$, where practically we can incrementally estimate $\mmu$ and subtract it from the stream observations. By relying on a two-timescale type approach~\citep{borkar1997stochastic} we can theoretically treat $\mmu$ as fixed when considering the convergence of our proposed method.
%
Similar to~\cite{pavlidis2016JMLR}, we focus on minimising the objective
\begin{align}\label{eq:objective}
    O(\v, b) := I_{f_X}(\v, b) + C\left(|b-\v^\top\mmu| - \alpha\right)_+^2,
\end{align}
over all unit vectors $\v$, where $\mmu = E[X]$ is the mean of the distribution. Here $C$ and $\alpha$ are chosen non-negative constants and $(x)_+ := \max\{0, x\}$. 
%
%\textcolor{red}{While we investiage the case of a fixed $\alpha$, it is easy to modify the analysis to allow an appropriate $\alpha$ to be estimated incrementally, again relying on a two-timescale approach.}
%
The purpose of penalising solutions for which $|b - \v^\top\mmu| > \alpha$ is to mitigate the possibility that the hyperplane simply passes through the tail of the distribution, thereby not separating high density regions. This penalty limits the distance of the hyperplane from the mean of the underlying distribution. While it is obviously true that distributions can be constructed for which the mean lies far away from the modes of the distribution, and in these highly skewed and long tailed examples this particular penalty is ineffective, we have found this approach to be very effective in many practical scenarios. For brevity going forward we will assume that $\mmu = \mathbf{0}$, where in practice we estimate $\mmu$ incrementally and subtract it from the sequence of observations, as in~\cite{CCIPCA}. By using a two-timescale approach~\citep{borkar1997stochastic} we can effectively ignore the variation in the estimated mean provided the sample mean converges, sufficient conditions for which are fairly weak.

Towards analysing the determination of minima of~(\ref{eq:objective}), we use the following set-up. First, let $\phi(x) = (2\pi)^{-1/2}\exp(-x^2/2)$ be the standard univariate Gaussian density evaluated at $x$. Then, for an initial estimate of the pair of parameters of the optimal hyperplane, say $(\v^{(0)}, b^{(0)})$, we use the update rule given by
\begin{align}
    \v^{(t+1)} &= \frac{1}{||\v^{(t)} - \gamma_1^{(t+1)}\u^{(t+1)}||}(\v^{(t)} - \gamma_1^{(t+1)}\u^{(t+1)}),\label{eq:vupdate}\\
    \u^{(t+1)} &:= \frac{b^{(t)} - \v^{(t)\top} X^{(t+1)}}{(h^{(t+1)})^3}\phi\left(\frac{\v^{(t)\top} X^{(t+1)} - b^{(t)}}{h^{(t+1)}}\right)X^{(t+1)},\label{eq:udef}\\
    b^{(t+1)} &= b^{(t)} + \gamma_2^{(t+1)}\left(\beta^{(t+1)} - 2 C (|b^{(t)}| - \alpha)_+\mathrm{sign}(b^{(t)})\right),\label{eq:bupdate}\\
    \beta^{(t+1)} &:= \frac{b^{(t)} - \v^{(t)\top} X^{(t+1)}}{(h^{(t+1)})^3}\phi\left(\frac{\v^{(t)\top} X^{(t+1)} - b^{(t)}}{h^{(t+1)}}\right),\label{eq:betadef}
\end{align}
where $h^{(1)}, h^{(2)}, ...$ is a scalar sequence of strictly positive bandwidth parameters, and $\gamma_1^{(1)}, \gamma_1^{(2)}, ...$ and $\gamma_2^{(1)}, \gamma_2^{(2)}, ...$ are deterministic scalar sequences of learning rates. In particular we use $h^{(t)} = s^{(t-1)}t^{-q}, \gamma_1^{(t)} = \bar\gamma_1 t^{-r}, \gamma_2^{(t)} = \bar\gamma_2 t^{-r}$ for each $t$, where $q, r, \bar\gamma_1, \bar\gamma_2 > 0$ are all fixed and $s^{(0)}, s^{(1)}, ...$ is a sequence which is almost surely bounded both above and away from zero below.

In the remainder of this section we investigate the sequence $(\v^{(0)}, b^{(0)}), (\v^{(1)}, b^{(1)}), ...$ in relation to the stationary points of $O(\v, b)$ under the constraint that $||\v|| = 1$. Note that for $(\v, b)$ to be such a stationary point it is sufficient that $\frac{\partial}{\partial b} O(\v, b) = 0$ and $\nabla_{\v} O(\v, b)^\top \left(\I - \v\v^\top\right) = \nabla_{\v} f_{\v^\top X}(b)^\top \left(\I - \v\v^\top\right) =  \mathbf{0}$, i.e., that the partial derivative of the objective w.r.t. $b$ is zero, and the partial gradient w.r.t. $\v$ is a zero except in the direction of $\v$. What we will show is that, for all practical purposes, the sequence converges in probability to a such stationary point. Formally we show that under mild conditions
\begin{align}
    \sum_{t=0}^{\infty}t^{-r}E\left[\left\|(\I - \v^{(t)}\v^{(t)\top})\nabla_\v f_{\v^{(t)\top} X}(\v^{(t)})\right\|^2\right] &< \infty,\\
    \sum_{t=0}^{\infty}t^{-r}E\left[\frac{\partial}{\partial b}O(\v^{(t)}, b^{(t)})^2\right] &< \infty.
\end{align}
While setting $0 < r \leq 1$ does not directly ensure that 
\begin{align*}
    \lim_{t\to\infty}E\left[\left\|(\I - \v^{(t)}\v^{(t)\top})\nabla_\v f_{\v^{(t)\top} X}(\v^{(t)})\right\|^2\right] = 0,\\
    \lim_{t\to\infty}E\left[\frac{\partial}{\partial b}O(\v^{(t)}, b^{(t)})^2\right] = 0,
\end{align*}
it is also stronger than simply ensuring the limit infima of these sequences are zero. It effectively ensures that, for all $\epsilon > 0$, the proportion of indices $t$ for which these expectations are greater than $\epsilon$ is zero in the limit. What this implies is that if we define, 
for each $t \in \mathbb{N}$, the discrete uniform random variable on $\{t, t+1, ..., 2t\}$ to be $N_t$, then, for all $\epsilon > 0$ we have
\begin{align*}
    \lim_{t\to\infty} P\left(\left\|\nabla_{\v} f_{\v^{(N_t)\top} X}(b^{(N_t)})^\top \left(\I - \v^{(N_t)}\v^{(N_t)\top}\right)\right\| > \epsilon\right) &= 0,\\
    \lim_{t\to\infty} P\left(\left|\frac{\partial}{\partial b} O(\v^{(N_t)}, b^{(N_t)})\right| > \epsilon\right) &= 0.
\end{align*}
For all practical purposes this is equivalent to convergence in probability, unless the time at which the incremental learning is terminated is very specific, or even ``unlucky''.

As far as we are aware existing studies of the convergence of SGD in problems using smoothing kernels have only focused on the fixed data set and bandwidth scenarios. In this case the stochastic gradients are unbiased since they estimate the gradient of the fixed bandwidth smoothing of the empirical distribution, and convergence to minima with respect to the convolution density can be achieved. In our case convergence to minima with respect to the true density is shown. However, should convergence of the type described above be preferred, then we can apply the same convergence established in this section to a sequence defined by resampling uniformly from the fixed data set at each time point, and adding small Gaussian perturbations with standard deviation equal to the fixed bandwidth used in these existing contexts.

Now, to facilitate our analysis, we make the following assumptions on the underlying density:
\begin{enumerate}
    \item For all $b_1, b_2 \in \R$ and $\v_1, \v_2 \in \R^{d}$, both with norm 1, we have
    \begin{align}
        \nonumber
        f_{\v_2\top X}(b_2) - &f_{\v_1^\top X}(b_1) \leq \nabla_{(\v, b)}f_{\v_1^\top X}(b_1)^\top((\v_1, b_1) - (\v_2, b_2))\\
        & + L(||\v_1-\v_2||^2 + (b_1-b_2)^2 + |b_1-b_2|\cdot||\v_1-\v_2||),\label{eq:diffbound}
    \end{align}
    for some constant $L$.
    \item The random variable $X$ has finite second moment, i.e., $E[||X||^2] < \infty$.
    \item For all $\v \in \R^d$ with $||\v|| = 1$, all of $f_{\v^\top X}(b), f_{\v^\top X}^\prime(b)$ and $f_{\v^\top X}^{\prime\prime}(b)$ are bounded.
    \item There exist $M, K, \eta > 0$, with $\eta$ small (practically 0.2 or smaller) such that for all $\v; ||\v|| = 1,$ we have $|f_{\v^\top X}^{\prime\prime\prime}(b)| < K|b|^{-\eta}$ for all $|b| \geq M$.
\end{enumerate}
While it is not always easy to verify Assumption 1, we show that this condition is held by finite Gaussian mixtures. Details are given in the appendix. This is an important class of distributions since all continuous distributions can be approximated arbitrarily well by one in this class. Assumptions 2--4 are fairly standard, and also clearly hold for finite Gaussian mixtures since if $X$ has a finite Gaussian mixture density then $\v^\top X$ has a (univariate) finite Gaussian mixture density. Importantly, as can be seen in the discussion in the appendix, the boundedness of $f_{\v^\top X}^\prime(b)$ ensures the boundedness of $\nabla_{\v}f_{\v^\top X}(b)$.

Now, towards establishing the convergence of the proposed update scheme, first note that it is straightforward to verify that for all $a, b \in 
\R$ we have $(|a|-\alpha)_+^2 - (|b|-\alpha)_+^2 \leq 2(|b|-\alpha)_+\mathrm{sign}(b)(a-b) + (a-b)^2$. Combining this with Assumption 1 we therefore have, for each $t$,
\begin{align}\nonumber
    O(\v^{(t+1)}, &b^{(t+1)}) - O(\v^{(t)}, b^{(t)}) \leq  \nabla_{(\v, b)}O(\v^{(t)}, b^{(t)})^\top\left((\v^{(t+1)}, b^{(t+1)}) - (\v^{(t)}, b^{(t)})\right)\\
    \nonumber
    &+ (L + C)\left(||(\v^{(t+1)}, b^{(t+1)}) - (\v^{(t)}, b^{(t)})||^2 + |b^{(t+1)}-b^{(t)}|\cdot ||\v^{(t+1)} - \v^{(t)}||\right)\\
    \nonumber
    =& \nabla_{\v}f_{\v^{(t)\top}X}(b^{(t)})^\top(\v^{(t+1)} - \v^{(t)}) + \frac{\partial}{\partial b}O(\v^{(t)}, b^{(t)})(b^{(t+1)} - b^{(t)})\\
    &+ (L + C)\left(||(\v^{(t+1)}, b^{(t+1)}) - (\v^{(t)}, b^{(t)})||^2 + |b^{(t+1)}-b^{(t)}|\cdot ||\v^{(t+1)} - \v^{(t)}||\right).\label{eq:conv_ineq}
\end{align}
%

%%%%%%%%%%%%%%%%%%%%%%%%%%%%%%%%%%%%%%%%%%%%%%%%%%%%%%%%%%%%%%%%
\iffalse

The function $\Pi(\cdot, \cdot)$ in the above is given by the following transformation\footnote{we refer to $\Pi$ as a transformation rather than a projection as it is not necessarily the case that it outputs the nearest pair which satisfies the constraints on $\v$ and $b$.},
%
\begin{align*}
    \Pi(\v, b) &= \left(\tilde \v, \tilde b\right)\\
    \tilde \v &:= \frac{1}{ ||\v||}\v,\\
    \tilde b &:= \max\left\{\min\left\{b, \tilde \v^\top \boldsymbol{\mu} + \alpha\sqrt{\tilde \v^\top\Sigma\tilde\v}\right\},  \tilde \v^\top \boldsymbol{\mu} - \alpha\sqrt{\tilde \v^\top\Sigma\tilde\v} \right\},
\end{align*}
%
where $\boldsymbol{\mu}$ and $\Sigma$ are the mean and covariance of $X^{(1)}$, respectively, and $\alpha > 0$ is some chosen parameter. This transformation constrains $\v$ to be inside the unit ball and restricts $b$ from diverging towards $\pm \infty$. The purpose of constraining the value of $b$ is to mitigate the possibility that the solution will only pass through the tail of $F(\cdot)$, rather than passing between clusters as is desired. The smaller the value of $\alpha$, the more tightly constrained is the optimal hyperplane to pass close to the mean $\boldsymbol{\mu}$. Note that while practically the first two moments of the underlying distribution must be learned from the data as well, theoretically they can be treated as fixed by using a two-timescale approach~\citep{borkar1997stochastic}.

\fi
%%%%%%%%%%%%%%%%%%%%%%%%%%%%%%%%%%%%%%%%%%%%%%%%%%%%%%%%%%%%%%%%%%%%%%%%%%%%%%
 \noindent
Now, it is relatively straightforward to show that
%for large $t$ we have
%
\begin{align*}
    \bigg\|\v^{(t)} - \gamma_1^{(t+1)} \u^{(t+1)}\bigg\|^{-1} = 1 + \gamma_1^{(t+1)}\v^{(t)\top}\u^{(t+1)} + \mathcal{O}\left((\gamma_1^{(t+1)}\v^{(t)\top}\u^{(t+1)})^2\right),
\end{align*}
where $\u^{(t+1)}$ is the stochastic estimate of the gradient of the objective w.r.t. $\v$ at time $t+1$, introduced in Eq.~(\ref{eq:udef}). We therefore find that
\begin{align*}
    \v^{(t+1)} - \v^{(t)} =& -\gamma_1^{(t+1)}\bigg(\u^{(t+1)} - \v^{(t)\top}\u^{(t+1)}\v^{(t)}\bigg)+ \mathcal{O}\left((\gamma_1^{(t+1)}\v^{(t)\top}\u^{(t+1)})^2\right)\\
    =& -\gamma_1^{(t+1)}\left(\I - \v^{(t)}\v^{(t)\top}\right)\u^{(t+1)} + \mathcal{O}\left((\gamma_1^{(t+1)}\v^{(t)\top}\u^{(t+1)})^2\right).
\end{align*}
This give us
\begin{align*}
    \nabla_{\v}f_{\v^{(t)\top}X}(b^{(t)})^\top(\v^{(t+1)} - \v^{(t)}) =& -\gamma_1^{(t+1)}\nabla_{\v}f_{\v^{(t)\top}X}(b^{(t)})^\top\left(\I - \v^{(t)}\v^{(t)\top}\right)\u^{(t+1)} + o(\gamma_1^{(t+1)}\v^{(t)\top}\u^{(t+1)})\\
    =& -\gamma_1^{(t+1)}\nabla_{\v}f_{\v^{(t)\top}X}(b^{(t)})^\top\left(\I - \v^{(t)}\v^{(t)\top}\right)\nabla_{\v}f_{\v^{(t)\top}X}(b^{(t)})\\
    &-\gamma_1^{(t+1)}\nabla_{\v}f_{\v^{(t)\top}X}(b^{(t)})^\top\left(\I - \v^{(t)}\v^{(t)\top}\right)\epsilon_{\v}^{(t+1)}\\
    & + \mathcal{O}\left((\gamma_1^{(t+1)}\v^{(t)\top}\u^{(t+1)})^2\right),
\end{align*}
where $\epsilon_{\v}^{(t+1)} := \u^{(t+1)} - \nabla_{\v} f_{\v^{(t)^\top} X}(b^{(t)})$ is the error of the $(t+1)$-th stochastic gradient w.r.t. $\v$, and since $\nabla_{\v}f_{\v^{\top}X}(b)$ is bounded.
In addition, consider that
\begin{align*}
    \frac{\partial}{\partial b}O(\v^{(t)}, b^{(t)})(b^{(t+1)} - b^{(t)}) &= -\gamma_2^{(t+1)}\frac{\partial}{\partial b}O(\v^{(t)}, b^{(t)})^2 - \gamma_2^{(t+1)}\frac{\partial}{\partial b}O(\v^{(t)}, b^{(t)})\epsilon_b^{(t+1)},
\end{align*}
where $\epsilon_b^{(t+1)}$ is the error of the estimate of the partial derivative of $f_{\v^\top X}(b)$ with respect to $b$ at time $(t+1)$.
In all, we therefore find, after rearranging Eq.~(\ref{eq:conv_ineq}), that
\begin{align}\nonumber
    \gamma_1^{(t+1)}&\nabla_\v f_{\v^{(t)\top} X}(b^{(t)})^\top\left(\I - \v^{(t)}\v^{(t)\top}\right)\nabla_\v f_{\v^{(t)\top} X}(b^{(t)}) + \gamma_2^{(t+1)}\frac{\partial}{\partial b}O(\v^{(t)}, b^{(t)})^2 \leq \\
    \nonumber
    & O(\v^{(t)}, b^{(t)}) - O(\v^{(t+1)}, b^{(t+1)}) - \gamma_1^{(t+1)}\nabla_\v f_{\v^{(t)\top} X}(b^{(t)})^\top\left(\I - \v^{(t)}\v^{(t)\top}\right)\epsilon_\v^{(t+1)}\\
    \nonumber
    & - \gamma_2^{(t+1)}\frac{\partial}{\partial b} O(\v^{(t)}, b^{(t)}) \epsilon_b^{(t+1)}\\
    \nonumber
    &+ (L + C)\left(||\v^{(t+1)}-\v^{(t)}||^2 + (b^{(t+1)}-b^{(t)})^2 + |b^{(t+1)}-b^{(t)}|\cdot ||\v^{(t+1)}-\v^{(t)}||\right)\\
    &+ \mathcal{O}\left((\gamma_1^{(t+1)}\v^{(t)\top}\u^{(t+1)})^2\right).\label{eq:conv_ineq2}
\end{align}
Now, it can be shown that if we let $\X^{(1:t)} = X^{(1)}, ..., X^{(t)},$ then
\begin{align}
    E\left[ \epsilon_\v^{(t+1)}| \X^{(1:t)}\right] &= \mathcal{O}(t^{-2q}),\label{eq:eps_error}\\
    E[||\u^{(t+1)}||^2|\X^{(1:t)}] &= \mathcal{O}(t^{2q})\label{eq:u^2_error}\\
    E\left[ \frac{\partial}{\partial b}O(\v^{(t)}, b^{(t)})\epsilon_b^{(t+1)}\right] &= \mathcal{O}(t^{-\eta} + t^{\eta - 2q}),\label{eq:dbeps_error}
\end{align}
where $q, \eta > 0$ were introduced in relation to the sequence of bandwidth parameters, and in Assumption 4, respectively, and with the final equation holding for $t$ large enough that $t^{\eta} > M$, for $M$ also given in Assumption 4. We leave details of the associated derivations to the appendix. We therefore have,
\begin{align*}
    E\left[\gamma_1^{(t+1)}\nabla_\v f_{\v^{(t)\top} X}(b^{(t)})^\top\left(\I - \v^{(t)}\v^{(t)\top}\right)\epsilon_\v^{(t+1)}\right] &= \mathcal{O}(t^{-(r+2q)}),\\
    E\left[\gamma_2^{(t+1)}\frac{\partial}{\partial b}O(\v^{(t)}, b^{(t)})\epsilon_b^{(t+1)}\right] &= \mathcal{O}(t^{-(r+\eta)}+t^{\eta-(r+2q)}),\\
    E[(\gamma_1^{(t+1)}\v^{(t)\top} \u^{(t+1)})^2] &= \mathcal{O}(t^{-2(r-q)}),
\end{align*}
where $r$ was introduced in relation to the definitions of $\gamma_1^{(t)}, \gamma_2^{(t)}$. The above arise using the law of total expectation and the fact that $(\v^{(t)}, b^{(t)})$ is fully determined by $\X^{(1:t)}$ and $(\v^{(0)}, b^{(0)})$.

We can also show (once more, details are given in the appendix) that
\begin{align}
    E\left[||\v^{(t+1)} - \v^{(t)}||^2 | \X^{(1:t)}\right] &= \mathcal{O}(t^{-2(r-2q)}),\label{eq:vmv_error}\\
    E\left[(b^{(t+1)} - b^{(t)})^2| \X^{(1:t)}\right] &= \mathcal{O}(t^{-2(r-2q)}),\label{eq:bmb_error}\\
    E\left[|b^{(t+1)} - b^{(t)})| \cdot ||\v^{(t+1)} - \v^{(t)}|| \ | \X^{(1:t)}\right] &= \mathcal{O}(t^{-2(r-2q)}),\label{eq:bmbvmv_error}
\end{align}
with the second holding for $t \geq (2\bar\gamma_2C)^{1/r}$.
%, \textcolor{red}{ and where the constants in the right hand sides are independent of $\X^{(1:t)}$.}
By taking the expectation on both sides of Eq.~(\ref{eq:conv_ineq2}), and summing over $t = 0, ..., T-1$ we obtain
\begin{align}
    \nonumber
    \sum_{t=0}^{T-1}&\gamma_1^{(t+1)}E\left[\left\|(\I - \v^{(t)}\v^{(t)\top})\nabla_\v f_{\v^{(t)\top} X}(\v^{(t)})\right\|^2\right] + \sum_{t=0}^{T-1}\gamma_2^{(t+1)}E\left[\frac{\partial}{\partial b}O(\v^{(t)}, b^{(t)})^2\right] \leq\\
    \nonumber
    & O(\v^{(0)}, b^{(0)}) - O(\v^{(T)}, b^{(T)}) + D\sum_{t=0}^{T-1} (t^{-(r+2q)} + t^{-(r+\eta)} + t^{-2(r-q)} + t^{-2(r-\eta/2)})\\
    \leq& O(\v^{(0)}, b^{(0)}) + D\sum_{t=0}^{T-1} (t^{-(r+2q)} + t^{-(r+\eta)} + t^{-2(r-q)} + t^{-2(r-\eta/2)}),
\end{align}
for some constant $D>0$ independent of $T$, where the second inequality comes from the fact that $O(\v, b) \geq 0$ for all $(\v, b)$. Now, if we choose $q, r$ for which $0 < r \leq 1, r + 2q > 1, r-q > 0.5, r + \eta > 1, q \geq \eta,$ and $r-\eta/2 > 0.5$, where $\eta$ is given in Assumption 4, then the right hand side is finite in the limit $T\to \infty$, and hence
\begin{align}
    \sum_{t=0}^{\infty}\gamma_1^{(t+1)}E\left[\left\|(\I - \v^{(t)}\v^{(t)\top})\nabla_\v f_{\v^{(t)\top} X}(\v^{(t)})\right\|^2\right] &< \infty,\\
    \sum_{t=0}^{\infty}\gamma_2^{(t+1)}E\left[\frac{\partial}{\partial b}O(\v^{(t)}, b^{(t)})^2\right] &< \infty,
\end{align}
as required.

\iffalse

\subsection{Obtaining Multiple Non-Redundant Low Density Hyperplanes}

A simple approach for ensuring that multiple hyperplanes, say $H(\v_1, b_1), ..., H(\v_m, b_m)$, are not merely finding the same minimiser of the objective in Eq.~(\ref{eq:objective}) is to encourage the vectors $\v_1, ..., \v_m$ to be (approximately) orthogonal. This can be achieved simply as follows. When performing an update for $\v_i, i > 1$, first project the observation into the null space of $\v_1, ..., \v_{i-1}$. Given some initial estimates for a collection of minimum density hyperplanes, parameterised by $(\v_1^{(0)}, b_1^{(0)}), ..., (\v_m^{(0)}, b_m^{(0)})$, a complete set of steps for updating these with the receipt of $X^{(t+1)}$ is given by
%
\begin{enumerate}
    \item[-] For $i = 1, ..., m$ do:
    \begin{enumerate}
        \item[-] if $i=1$ do: $\x \gets X^{(t+1)}$
        \item[-] else do: $\x \gets \x - (\v_{i-1}^{(t+1)\top} \x) \v_{i-1}^{(t+1)}$ 
        \item[-] $\u \gets \frac{b_i^{(t)} - \v_i^{(t)\top} \x}{(h^{(t+1)})^3}\phi\left(\frac{\v_i^{(t)\top} \x - b_i^{(t)}}{h^{(t+1)}}\right)\x$
        \item[-] $\v_i^{(t+1)} \gets  \frac{1}{||\v_i^{(t)} - \gamma_1^{(t+1)}\u||}(\v_i^{(t)} - \gamma_1^{(t+1)}\u)$
        \item[-] $b_i^{(t+1)} \gets b_i^{(t)} + \gamma_2^{(t+1)}\left(\frac{b_i^{(t)} - \v_i^{(t)\top} \x}{(h^{(t+1)})^3}\phi\left(\frac{\v_i^{(t)\top} \x - b_i^{(t)}}{h^{(t+1)}}\right) - 2 C (|b_i^{(t)}| - \alpha)_+\mathrm{sign}(b_i^{(t)})\right)$.
    \end{enumerate}
\end{enumerate}
%

\fi

\section{Clustering with Low Density Hyperplanes}\label{sec:clustering}

\begin{figure}[h]
    \centering
    \includegraphics[width = 13cm, height = 10cm]{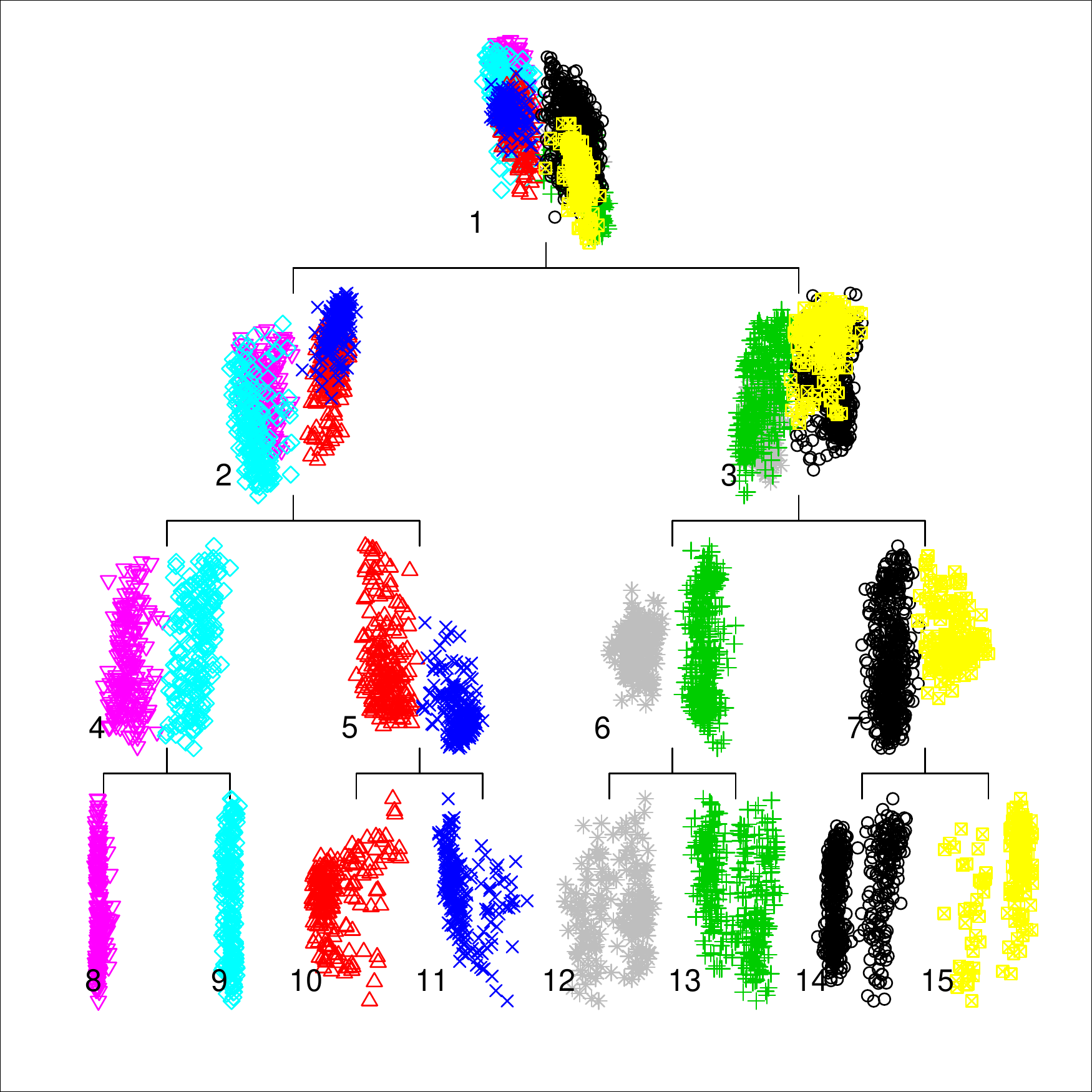}
    \caption{A divisive hierarchical clustering model in which the partitions at each node in the hierarchy are formed by a hyperplane. Colours and point characters indicate the assignments from the complete model into eight clusters.}
    \label{fig:treeplot}
\end{figure}

A common approach for clustering with hyperplanes is within a divisive hierarchical model. Here all the data are assigned to the highest level in the hierarchy (the root node). They are then split in two by a hyperplane, and the two resulting subsets are allocated respectively to the two nodes at the second level in the hierarchy (the child nodes of the root node). Each of these subsets is then split in two by a hyperplane, and their subsets passed to the next level, etc. This divisive approach continues until a certain depth is reached. A simple indexing strategy for the nodes in such a model is to allocate index 1 to the root node, and to the child nodes of node $i$ (those to which it passes) allocate the indices $2i$ and $2i+1$. Figure~\ref{fig:treeplot} shows an example of such a model. The numbers indicate the indices of the different nodes, with a total of $2^{D} - 1$, where $D$ is the depth of the model. The different colours and point characters indicate the partition into eight clusters, which are formed by the subsets of points allocated to the different nodes at the lowest level in the hierarchy. The figure also illustrates one of the benefits of such models, as the partitions at internal nodes, as well as the potential partitions at terminal nodes, can be visualised, offering some interpretability in the model as well as being potentially useful for subjective validation~\citep{hofmeyrPPCI}. For example, in this instance there is some evidence that further partitioning in nodes 12, 13, 14 and 15 is reasonable, since there is evidence of multiple clusters within the observations assigned to those nodes. While it is not always straightforward to find a low-dimensional representation of the observations which displays their cluster structure, by design the vectors parameterising the hyperplanes at each node are ones orthogonal to which there will tend to be a low density separator for the observations assigned there. The horizontal axes in the figure correspond with the projections of these observations onto these vectors, while the vertical axes can be chosen arbitrarily since the prevailing cluster structure is visible along the horizontal direction.

Within the incremental estimation framework, suppose that we have the current set of hyperplanes $(\v_1, b_1), ..., (\v_{k}, b_{k})$ for the different nodes in the hierarchy, as well as estimates of the means of the subsets of the sequence of i.i.d. random variables, $X^{(1)}, X^{(2)}, ...$, which are allocated to each node, say $\hmmu_{1}, ..., \hmmu_{k}$. We also have the numbers of observations which have so far been allocated to each of the nodes, $t_1, ..., t_k$.
%
%, as well as the assignments for the first $t$ observations in the sequence, stored as the set $\mathcal{C}$.
%
Algorithm~\ref{alg:incremental_update} contains pseudo-code describing the steps taken to update these objects with the receipt of the next observation, say $X^{(t+1)}$. In the pseudo-code ``id'' is used to represent the index of the node which is currently being updated.
%
%, and $c$ is the current assignment of observation $X^{(t+1)}$.
%
The vector $\u$ and scalar $\beta$ are used to store the stochastic gradients.
%
%Note that in this description the allocation of points to clusters is done simultaneously with the estimation of the model. Early on in the estimation the accuracy of the model will not be high, and so it may be beneficial to store an initial set of observations and only allocate cluster labels once the model has had an opportunity to begin to converge. Alternatively the cluster allocations can be done seperately as is common in the online/offline format of most data stream clustering algorithms~\citep{...}.
%

\begin{algorithm}[h]
  %\algsetup{linenosize=\tiny}
  \footnotesize
\begin{algorithmic}
\STATE Input: $(\v_1, b_1), ..., (\v_{k}, b_{k}), \hmmu_{1}, ..., \hmmu_{k}, t_1, ..., t_k$ and $X^{(t+1)}$
\STATE id $\gets 1$ %\hfill {\em\# set index to highest/root node\\}
\WHILE{id $ < k$}
%\STATE $c \gets$ id
\STATE $t_{\mathrm{id}} \gets t_{\mathrm{id}}+1$
\STATE $\hmmu_{\mathrm{id}} \gets \frac{t_{\mathrm{id}}-1}{t_{\mathrm{id}}} \hmmu_{\mathrm{id}} + \frac{1}{t_{\mathrm{id}}}X^{(t+1)}$
\STATE $\u \gets \frac{b_{\mathrm{id}} - \v_{\mathrm{id}}^\top (X^{(t+1)} - \hmmu_{\mathrm{id}})}{(h^{(t_{\mathrm{id}})})^3} \phi\left(\frac{b_{\mathrm{id}} - \v_{\mathrm{id}}^\top (X^{(t+1)} - \hmmu_{\mathrm{id}})}{h^{(t_{\mathrm{id}})}}\right)(X^{(t+1)}-\hmmu_{\mathrm{id}})$
\STATE $\beta \gets 2C(|b_{\mathrm{id}}|-\alpha)_+\mathrm{sign}(b_{\mathrm{id}}) - \frac{b_{\mathrm{id}} - \v_{\mathrm{id}}^\top (X^{(t+1)} - \hmmu_{\mathrm{id}})}{(h^{(t_{\mathrm{id}})})^3} \phi\left(\frac{b_{\mathrm{id}} - \v_{\mathrm{id}}^\top (X^{(t+1)} - \hmmu_{\mathrm{id}})}{h^{(t_{\mathrm{id}})}}\right)$
\STATE $\v_{\mathrm{id}} \gets \frac{1}{||\v_{\mathrm{id}} - \gamma_1^{(t_{\mathrm{id}})}\u||}(\v_{\mathrm{id}} - \gamma_1^{(t_{\mathrm{id}})}\u)$
\STATE $b_{\mathrm{id}} \gets b_{\mathrm{id}} - \gamma_2^{(t_{\mathrm{id}})}\beta$
\IF{$\v_{\mathrm{id}}^\top (X^{(t+1)} - \hmmu_{\mathrm{id}}) < b_{\mathrm{id}}$}
{\STATE $\mathrm{id} \gets 2\mathrm{id}$}
\ELSE
{\STATE $\mathrm{id} \gets 2\mathrm{id}+1$}
\ENDIF
\ENDWHILE
%\STATE $\mathcal{C} \gets \mathcal{C} \cup \{c - 2^{d-1} + 1\}$
\end{algorithmic}
\caption{Incremental update of hierarchical model using minimum density hyperplanes \label{alg:incremental_update}}
\end{algorithm}

Now, it should be clear that a complete hierarchical model of a given depth will not always be appropriate, since the number of clusters is restricted to be a power of two. Furthermore, it is generally not realistic to be able to define an appropriate topology for the hierarchy {\em a priori}. In practice, therefore, we begin by estimating a complete model of a chosen depth, and then apply an efficient offline pruning algorithm to obtain a final clustering solution. The pruning objective we employ is based on the within-cluster sum-of-squares used in the classic $k$-means model. The pruning in fact only requires access to these sum-of-squares values for each of the subsets of the observations which have been allocated to the different nodes in the hierarchy. Since these can be estimated incrementally with the construction of the hierarchical model itself, this step does not require access to the observations themselves. This pruning is conducted greedily by repeatedly removing the split at the internal node which results in the least increase in the total sum-of-squares objective. Specifically, suppose that $SS(1), ..., SS(2^D-1)$ represent the sum-of-squares values from each of the nodes in the complete hierarchy, which we denote $H_{2^{D-1}}$ since it contains a total of $2^{D-1}$ clusters (terminal/leaf nodes). Now, for any pruned model obtained from $H_{2^{D-1}}$, say $H$, let $P(H)$ be the set of its {\em pre-terminal nodes}, i.e., those whose child nodes are both terminal nodes in $H$. The node at which we prune model $H$ is then
\begin{align*}
    \mathrm{argmin}_{j \in P(H)} SS(j) - \left(SS(2j) + SS(2j+1)\right).
\end{align*}
Through iterative pruning we obtain the sequence of hierarchical models $H_{2^{D-1}}, H_{2^{D-1}-1}, ..., H_{1}$. To choose from among these we use a simple elbow rule which selects the model containing $K$ clusters where $K$ minimises
\begin{align}
\nonumber
\arctan&\left(\frac{K-1}{K_{max}-1}\frac{SS(H_1)-SS(H_{K_{max}})}{SS(H_1)-SS(H_{K})}\right)\\
&+ \arctan\left(\frac{K_{max}-1}{K_{max}-K}\frac{SS(H_{K})-SS(H_{K_{max}})}{SS(H_1)-SS(H_{K_{max}})}\right),\label{eq:elbow}
\end{align}
where for a hierarchical model $H$ we have used $SS(H)$ to be the total sum-of-squares from its leaf nodes. This selects $K$ to minimise
the angle between the line segments joining the first to $K$-th and $K$-th to $K_{max}$-th points on the scaled graph of $K$ against total sum-of-squares. The distinction of $K_{max}$ from $2^{D-1}$ is deliberate since we would like for the final model to be minimally influenced by the selection of its tuning parameters, of which $D$ is one. In order to achieve this we compute the elbow in~(\ref{eq:elbow}) for a range of values for $K_{max}$ and select as the final $K$ that which occurs most frequently.

\section{Experimental Results}\label{sec:experiments}

In this section we report on the results from experiments conducted using the proposed method. For performance comparisons, we also report the results obtained using $k$-means models and another hierarchical clustering model based on low density separating hyperplanes~\citep[dePDDP]{tasoulis2010enhancing}. These methods are most relevant as they can both be implemented efficiently and have similar limitations to the proposed approach in the types of clusters which they can identify. Moreover, both have had incremental/online variants developed. SPDC~\citep{tasoulis2012clustering} is an online variant of dePDDP, while online estimation for $k$-means has numerous variants. We consider the BICO algorithm~\citep{BICO}, which is based on the data structures used by the influential BIRCH~\citep{birch} algorithm, and has specific theoretical guarantees related to the $k$-means solution(s). We use the standard implementation of $k$-means provided in {\tt R}~\citep{R} which is based on the algorithm of~\cite{hartiganWong}, and the implementation of BICO provided in the package {\tt stream}~\citep{streamCRAN}. We used our own implementations of dePDDP and SPDC.

\subsection{Clustering Accuracy and Run-time Benchmarking}

In order to obtain relevant performance comparisons for assessment we applied the clustering algorithms to a large collection of publicly available benchmark data sets. Most of these were obtained using the {\tt R} package {\tt pmlbr}~\citep{pmlbr}. To select data sets available in this package, we applied the clustering algorithms to all classification data sets containing at least 5000 observations. 
%
%We then removed those data sets on which none of the clustering algorithms achieved a normalised mutual information~\citep[NMI]{NMI} score above 0.05. While classification data sets are useful benchmarks for clustering since they allow for the comparison of the cluster assignments with the ground truth labels, very low NMI for all of the clustering algorithms suggests that the true class distribution does not correspond with how we interpret clusters, or that the types of clusters therein are not estimable by the types of models we consider. Comparing clustering results in such situations may lead to fairly arbitrary conclusions. 
%
This selection resulted in a total of 27 data sets. In addition we included three more examples of reasonably high dimensionality:
%
%fashion MNIST data set~\citep{fashionMNIST}
Yale faces dataset B (compressed to 40$\times 30$) and two data sets (isolet and smartphone) obtained from the UCI machine learning repository~\citep{UCI}. Table~\ref{tb:datasets} includes summaries of these data sets in terms of number of observations, $n$, number of variables, $d$, number of classes, $k$, and a measure of class imbalance defined by the variance of the class proportions~\citep{pmlbr}.

For the proposed approach we set the depth of the initial hierarchical model to 8 (and hence the maximum potential number of clusters is 256), the $t$-th element in the sequence of bandwidths to be $\hat \sigma_{\v^{(t)}}^{(t)}t^{-0.2}$, where $\hat \sigma_{\v^{(t)}}^{(t)}$ is an estimate of the standard deviation of the random variable $\v^{(t-1)\top}X$ after time $t-1$, which can easily be obtained incrementally from the sequence of projections obtained from projecting the observations onto the corresponding values for $\v^{(\cdot)}$. The value of $C$ in the objective function was set to 10, and $\alpha$ was dynamically adjusted using $0.1\hat \sigma_{\v^{(t)}}^{(t)}$. We report the results of a sensitivity study to assess the robustness of the approach in a later subsection. %\footnote{Note that while in the theory we treated $\alpha$ as being fixed, we can easily modify the theory using a two-timescale approach, allowing us to effectively ignore the variation in an estimated $\alpha$.}.

\subsubsection{Clustering Accuracy}

Table~\ref{tb:perf} reports the Normalised Mutual Information (NMI) and Adjusted Rand Index (ARI) scores\footnote{The values of NMI and ARI in the table are multiplied by 100 to reduce the total number of digits needed to represent the same number of significant figures} obtained using each of the algorithms. The scores are averages from 20 applications of the algorithms, where for the incremental methods, since these depend on the order in which the observations are presented, the randomness is induced by randomly ordering the data sets as well as any innate randomness such as initialisation,
%
%(iMDH has no innate randomness),
%
and in the case of $k$-means the randomness is based on the random initialisation of the cluster centroids. The dePDDP algorithm is fully deterministic. Incremental algorithms were given two passes over the data set each time; one in which to estimate the model and one to allocate the points to clusters. For the proposed iMDH method we include results where the number of clusters is estimated as described in the previous section, as well as the case where the true number of classes is given. For this latter case, denoted iMDH$_k$, the same pruning approach is applied as previously described, and the solution containing the given number of clusters is returned.

\renewcommand{\arraystretch}{.7}
\begin{table}[h]
    \centering
    \begin{tabular}{l|rrrr||l|rrrr}
data set&$n$&$d$&$k$&imb.&data set&$n$&$d$&$k$&imb.\\
\hline
churn&5000&20&2&0.51&mushroom&8124&22&2&0.00\\
waveform$_{21}$&5000&21&3&0.00&agaricus\_lepi.&8145&22&2&0.00\\
waveform$_{40}$&5000&40&3&0.00&smartphone&10929&561&12&0.07\\
phoneme&5404&5&2&0.17&pendigits&10992&16&10&0.00\\ 
page\_blocks&5473&10&5&0.76&nursery&12958&8&4&0.09\\
texture&5500&40&11&0.00&magic&19020&10&2&0.09\\
optdigits&5620&64&10&0.00&letter&20000&16&26&0.00\\
Yale&5850&1200&10&0.00&krkopt&28056&6&18&0.05\\
isolet&6238&627&26&0.00&adult&48842&14&2&0.27\\
satimage&6435&36&6&0.03&shuttle&58000&9&7&0.59\\
clean2&6598&168&2&0.48&mnist&70000&784&10&0.00\\
ann\_thyroid&7200&21&3&0.79&fars&100968&29&8&0.16\\
ring&7400&20&2&0.00&sleep&105908&13&5&0.15\\
twonorm&7400&20&2&0.00&kddcup&494020&41&23&0.38\\
    \end{tabular}
    \caption{Data sets used in accuracy and runtime benchmarking.}
    \label{tb:datasets}
\end{table}

\renewcommand{\arraystretch}{.6}

\setlength{\tabcolsep}{5pt}
\begin{table}[h]
    \centering
    \begin{tabular}{l||rr|rr|rr|rr|rr|rr|rr|rr|}
&\multicolumn{2}{c|}{iMDH}&\multicolumn{2}{c|}{iMDH$_k$}&\multicolumn{2}{c|}{BICO}&\multicolumn{2}{c|}{$k$-means}&\multicolumn{2}{c|}{SPDC}&\multicolumn{2}{c|}{dePDDP}\\
\hline
data set & NMI  & ARI & NMI  & ARI & NMI  & ARI & NMI  & ARI & NMI  & ARI & NMI  & ARI\\
\hline
churn&\bf 2.8&\bf 0.8&0.9&-2.2&\underline{1.8}&\bf -1.4&0.9&-2.1&0.0&0.0&0.0&0.0\\
waveform$_{21}$&\bf 42.5&25.1&38.6&28.0&36.7&25.6&36.5&25.4&35.7&26.4&41.1&\bf 30.0\\
waveform$_{40}$&38.3&25.9&38.3&27.6&36.3&25.2&36.5&25.2&37.3&\bf 27.7&\bf 42.8&\bf 28.5\\
phoneme&14.2&5.1&13.8&\bf 13.9&12.5&\bf 14.7&\bf 17.4&\bf 14.1&14.4&5.4&15.0&9.7\\
page\_blocks&18.4&5.7&7.8&4.4&18.7&19.8&16.3&9.6&\underline{20.9}&\bf 27.5&\bf 24.0&\bf 32.3\\
texture&\underline{62.3}&43.0&\bf 64.5&\bf 47.8&54.6&34.9&62.1&\bf 46.0&31.8&7.6&33.5&7.7\\
optdigits&\bf 65.5&\bf 52.3&\bf 64.7&\bf 52.6&19.6&6.6&\bf 63.8&\bf 50.9&4.8&0.8&0.0&0.0\\
Yale&\bf 82.5&\bf 64.0&\bf 84.5&\bf 68.6&66.3&45.4&71.8&54.4&11.4&2.1&13.7&0.4\\
isolet&67.8&38.7&\bf 70.0&\underline{45.0}&67.1&40.0&\bf 70.4&\bf 46.8&42.6&9.2&40.2&6.9\\
satimage&59.1&47.5&59.3&50.3&50.4&41.1&\bf 61.2&\bf 52.8&56.5&44.5&60.6&47.2\\
clean2&5.4&3.8&2.3&-2.6&2.6&-3.0&2.5&-0.9&8.5&2.3&\bf 17.7&\bf 4.4\\
ann\_thyroid&\bf 3.9&0.1&0.6&-1.4&\underline{1.9}&-0.4&2.4&-3.0&\bf 5.1&\bf 3.3&0.0&0.0\\
ring&10.1&8.1&11.2&14.1&9.2&5.8&\bf 25.5&\bf 26.3&0.7&0.0&0.0&0.0\\
twonorm&75.5&75.5&84.4&91.0&83.7&90.6&\underline{85.0}&\bf 91.6&\underline{83.5}&\bf 89.8&84.2&90.9\\
mushroom&\bf 44.0&\bf 24.5&23.6&\bf 28.8&8.2&3.6&11.8&9.9&39.2&12.1&31.1&0.4\\
agaricus\_lepiota&\bf 39.7&\bf 21.9&22.0&\bf 27.3&6.3&3.9&13.7&12.1&37.4&7.6&30.8&0.2\\
coil2000&1.0&0.3&1.0&1.1&0.3&-0.3&\bf 1.2&\bf 2.2&0.1&0.3&0.0&0.0\\
smartphone&55.8&\underline{36.9}&\bf 58.6&\bf 39.0&49.9&32.3&54.5&35.1&\underline{52.7}&28.5&56.9&30.1\\
pendigits&\bf 70.0&\bf 55.1&\underline{67.5}&50.3&64.4&48.8&\bf 68.6&\bf 55.7&61.2&41.9&62.6&37.0\\
nursery&7.7&\underline{3.0}&5.0&\bf 3.7&5.1&\bf 4&8.5&\bf 6.6&0.0&0.0&\bf 28.9&0.5\\
magic&10.2&5.2&0.2&0.6&\bf 13.4&\bf 11.6&0.2&0.6&0.3&0.0&0.0&0.0\\
letter&26.8&8.9&\underline{35.9}&12.8&34.8&13.7&\bf 37.1&\bf 15.0&0.7&0.1&0.0&0.0\\
krkopt&10.2&\bf 3.4&11.2&3.2&\bf 12.8&\bf 3.6&\bf 12.8&\bf 3.6&5.1&0.7&5.1&0.3\\
adult&9.2&\underline{2.1}&7.0&\bf 2.1&4.4&\bf 3.1&7.4&\bf 3&8.8&-1.2&\bf 10.9&-4.5\\
shuttle&\bf 45.7&18.7&\bf 46.6&\bf 35.4&2.6&0.3&42.9&22.5&36&\bf 31.9&39.6&\bf 33.3\\
connect\_4&\bf 0.5&\underline{0.1}&0.2&\bf 0.1&0.1&-0.1&0.1&\underline{0.0}&\bf 0.4&\bf 0.4&0.0&0.0\\
mnist&\bf 41.2&\underline{26.3}&\bf 43.3&\bf 30.7&1.4&0.0&\bf 42.2&\bf 30.7&19.8&3.0&24.9&2.8\\
fars&11.4&\bf 7.5&12.2&\bf 6.5&10.2&0.1&12.3&\bf 6.1&12.5&\bf 6.8&\bf 20.8&-1.5\\
sleep&\underline{21.5}&9.6&19.0&10.8&\bf 21.7&\bf 13.8&\bf 22.6&\bf 14.0&4.1&1.9&0.0&0.0\\
kddcup&\bf 83.1&\bf 91.6&80.4&90.1&18.9&11.2&74.5&85.8&3.2&2.5&81.3&85.7\\
    \end{tabular}
    \caption{Clustering performance based on normalised mutual information (NMI) and adjusted Rand index (ARI). Scores are multiplied by 100. Bold font indicates average not distinguishable from highest average performance based on a $t$-test with size 0.01, while underlines indicate indistinguishability from highest at 0.05 level.}
    \label{tb:perf}
\end{table}

Bold font in the table indicates that the average performance of the corresponding method was not significantly less than that of the highest average performance using a $t$-test with size 0.01. Underlined figures indicate that the method's average performance was not significantly lower than the highest when using a test size of 0.05.
%
%The table also includes, in the final row, the number of times each method achieved one of the highest performances, based on the results of these tests.
%
The proposed method is highly competitive with even the batch implementation of $k$-means, and substantially outperforms the other incremental algorithms except in very few cases.

\begin{figure}[h]
    \centering
    \includegraphics[width = .48\textwidth]{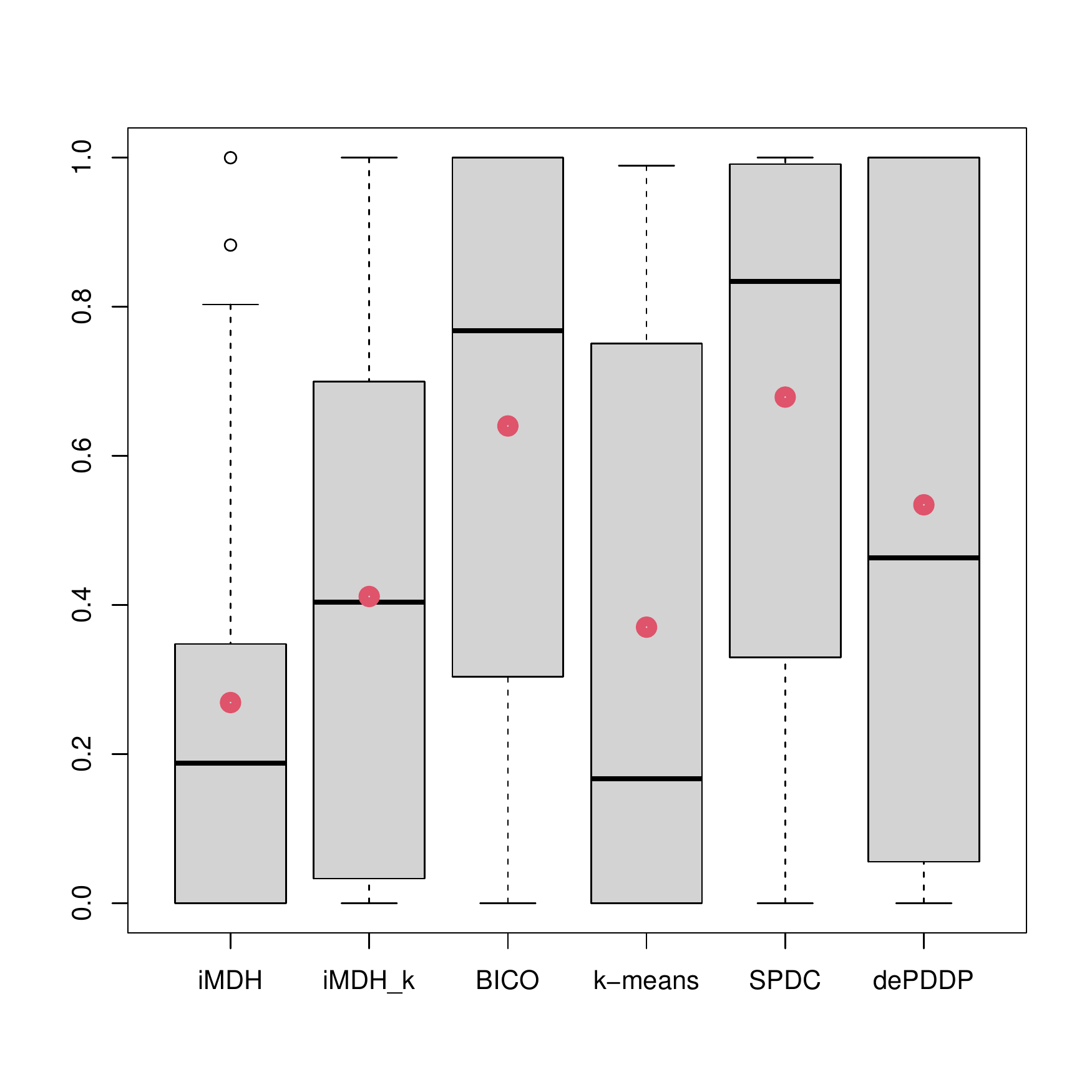}
    \includegraphics[width = .48\textwidth]{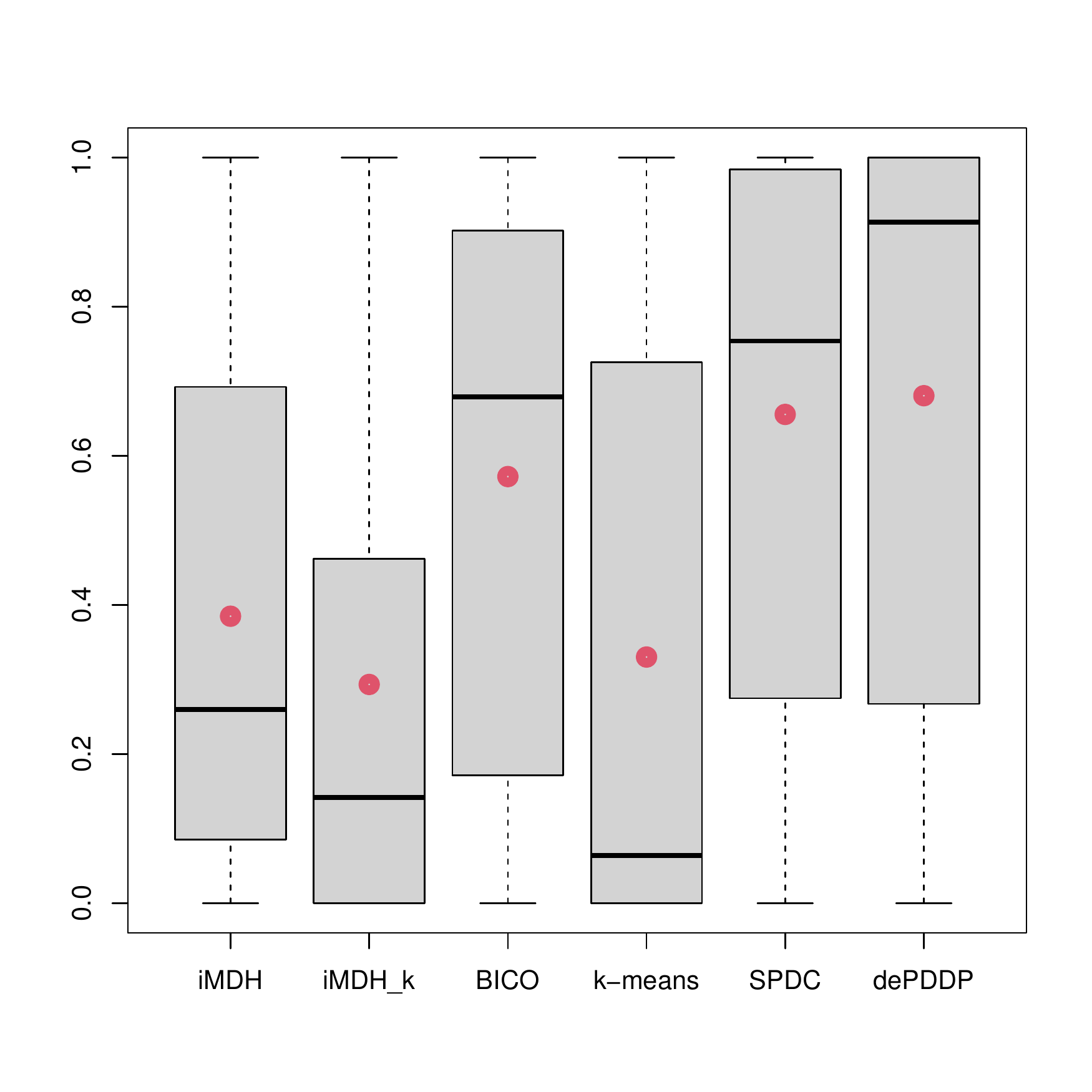}
    \caption{Boxplots of comparative performance based on normalised regret. In addition the means are indicated with red dots. Left: NMI. Right: ARI.}
    \label{fig:boxplots}
\end{figure}

The overall performance across the entire collection of data sets is summarised in Figure~\ref{fig:boxplots}. The figure includes boxplots of the {\em normalised regret} for each method, where the regret of a method on a given data set is the difference between the method's performance and the highest performance from all methods on that data set. The normalisation we apply simply transforms the regret values on each data set to span the interval $[0, 1]$. The purpose of normalisation is to standardise the regret values across data sets to make them more comparable with one another. The mean normalised regret in each case is also indicated with a red dot. These summaries corroborate the conclusions made from Table~\ref{tb:perf}, in that the overall performance of the proposed method is highly competitive with the batch version of $k$-means, and is substantially superior to that of the other incremental methods.

\subsubsection{Running time}

\begin{figure}[h]
    \centering
    \includegraphics[width = .9\textwidth]{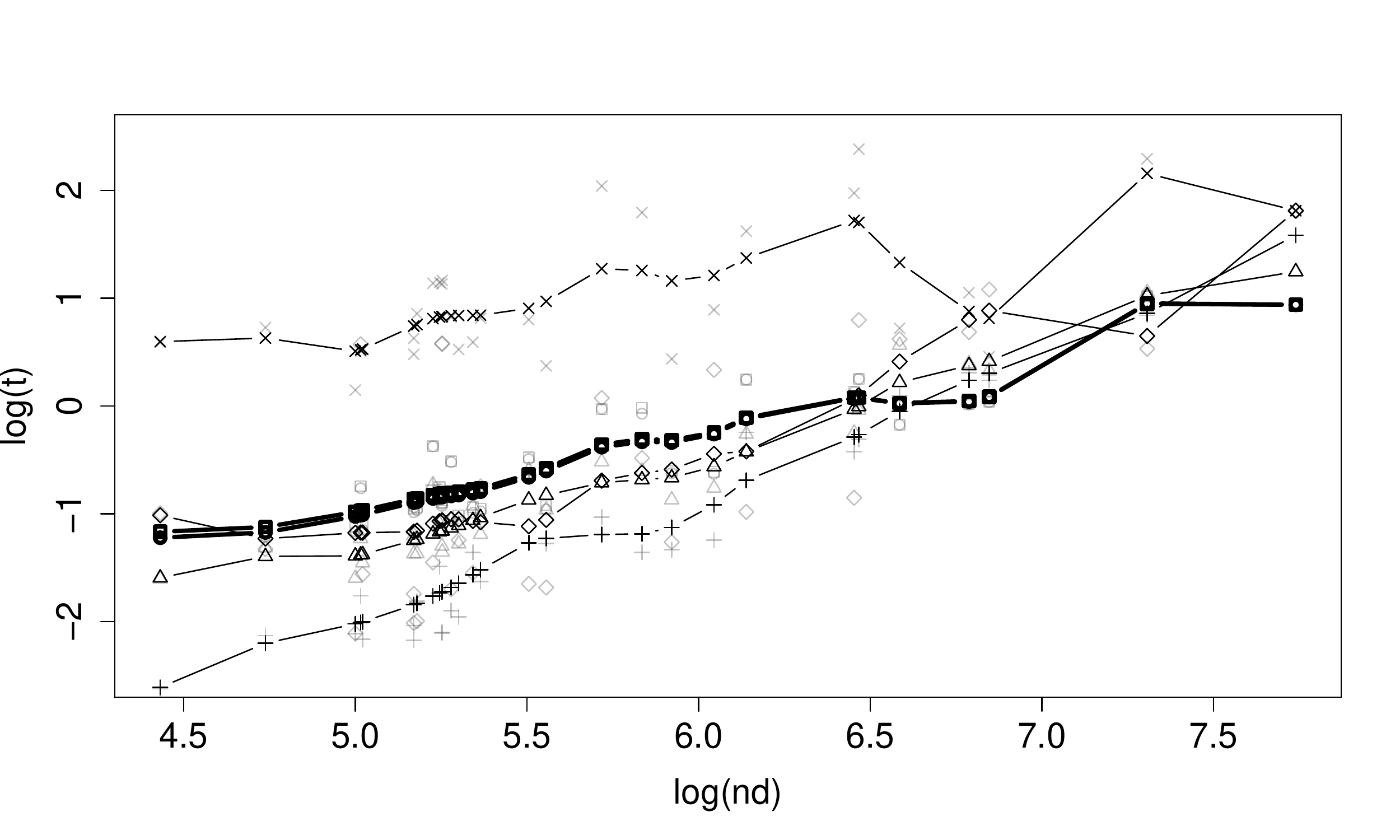}
    \caption{Plot of running time (in seconds) against $n \times d$ on a $\log_{10}$ scale. Average running times are shown in grey, while smoothed plots are shown in black. The plots for iMDH and iMDH$_k$ are highlighted in bold. iMDH ($\square$), iMDH$_k$ ($\circ$), BICO ($\triangle$), $k$-means ($+$), SPDC ($\times$), dePDDP ($\diamond$)}
    \label{fig:runtime}
\end{figure}

All of the methods considered have worst case running time which is linear in $n\times d$, except dePDDP which scales with $n\times d^2$ in the worst case. Figure~\ref{fig:runtime} shows plots of the average running times from 20 replications against $n \times d$, shown on a logarithmic scale for better interpretability. To further enable interpretation, the values from each method have been smoothed using a kernel smoother. The original values are shown in faint grey, while the smoothed curves are shown in black. The curves for iMDH and iMDH$_k$ have also been plotted in bold. As expected the running times of these are almost exactly the same as one another, since the only step excluded from iMDH$_k$ is the selection of $k$ using multiple elbow tests.

While for the smaller data sets the current implementation of iMDH is considerably slower than $k$-means (the fastest on the smaller data sets), its running time on the largest data sets is highly competitive and is lower than all other methods used for comparison on four data sets (smartphone, isolet, Yale and mnist)\footnote{after smoothing, the curves for iMDH and iMDH$_k$ lie above that of $k$-means for the isolet data set.}. It is worth noting that these four data sets all have relatively high dimensionality, all being more than 500 dimensional.

\subsection{Sensitivity Analysis}

In this section we provide a sensitivity analysis of the proposed method to some of its tuning parameters. We can separate these parameters into model parameters (tree depth; $\alpha$; and $C$) and learning parameters (the sequences $\{\gamma^{(t)}_1\}_{t=1}^\infty$; $\{\gamma^{(t)}_2\}_{t=1}^\infty$; and $\{h^{(t)}_1\}_{t=1}^\infty$). For the most part these have either intuitive or theoretical underpinnings which allow us to determine reasonable settings for consistent performance.

Setting the learning rates, $\gamma^{(t)}_1 = \sqrt{d}t^{-1}$ and  $\gamma^{(t)}_2 = t^{-1}$, aside from being consistent with the convergence analysis in Section~\ref{sec:convergence}, was dictated by the fact that the parameter $b$ can be seen as acting as a pivot for the hyperplane $H(\v, b)$ and, as a result, can have a very drastic effect on accurate learning of the optimal hyperplane if the sequence $b^{(0)}, b^{(1)}, ...$ is highly variable. As a result we prefer to have the learning of $b$ be slower than that of $\v$, to mitigate this variability (hence the factors $\sqrt{d}$ and 1 for the learning rates of the two components of $H(\v, b)$). Beyond this relationship the effect of different learning rates can be quite readily handled by varying only one of the sequences $\{\gamma^{(t)}_1\}_{t=1}^\infty$, $\{\gamma^{(t)}_2\}_{t=1}^\infty$, and $\{h^{(t)}_1\}_{t=1}^\infty$, because of a strong interplay between them. In particular, we choose to investigate the effect of the sequence of bandwidth parameters, as larger values for these smoothing parameters tend to lead to slower and more stable learning, whereas smaller values lead to faster but more highly variable learning.

The model parameters $\alpha$ and tree depth can both have a substantial influence on the performance as they dictate the flexibility of the estimation. In particular, a greater tree depth corresponds with a higher maximum number of clusters, and as a consequence a higher number of clusters in the pruned model as well. On the other hand small values of $\alpha$ tend to lead to more balanced cluster sizes, since the hyperplane at each node in the hierarchical model is constrained to lie closer to the mean of its observations. The parameter $C$ is arguably not a model parameter in the same sense as $\alpha$ and tree depth since its presence is only to affect the constraint on the distance of the hyperplane from the mean. While it can certainly influence the solution quality, it is not intended as a tuning parameter and rather we recommend simply leaving this parameter in its default setting.

\subsubsection{Varying Tree Depth}

The main effect of tree depth is that the greater the depth the greater the maximum number of clusters. A result of being able to select a model with more clusters is that there is an inflated possibility of erroneously overestimating an appropriate number of clusters. Notice that, how we have implemented the method, increasing the depth by one doubles the maximum number of clusters. As we will see in the results in this subsection, the number of selected clusters appears to grow closer to quadratically in the depth, before beginning to level out, rather than exponentially. Of greater importance, arguably, than the potential overestimation of $k$, however, is the fact that increasing the depth drastically increases the running time of the method. We considered tree depth between three (up to eight clusters) and twelve (up to 4096 clusters) to investigate the effect on model performance and running time. We report results from four of the datasets, namely optidigits ($n = 5620, d = 64, k = 10$); pendigits ($n = 10992, d = 16, k = 10$); isolet ($n = 6238, d = 627, k = 26$) and shuttle ($n = 58000, d = 9, k = 7$). The pendigits data set is included as one on which all methods perform reasonably well, while on optidigits the performance of SPDC and dePDDP is very poor compared with the others. Isolet is included as a high dimensional example for which the number of observations does not make estimating depth 12 trees computationally prohibitive. Shuttle is reasonably large in the number of observations, where on larger data sets the running time for depth twelve trees is prohibitive without availing additional insights.

\begin{figure}[h]
    \centering
    \subfigure[Optidigits]{\includegraphics[width = 0.45\textwidth, height = 6cm]{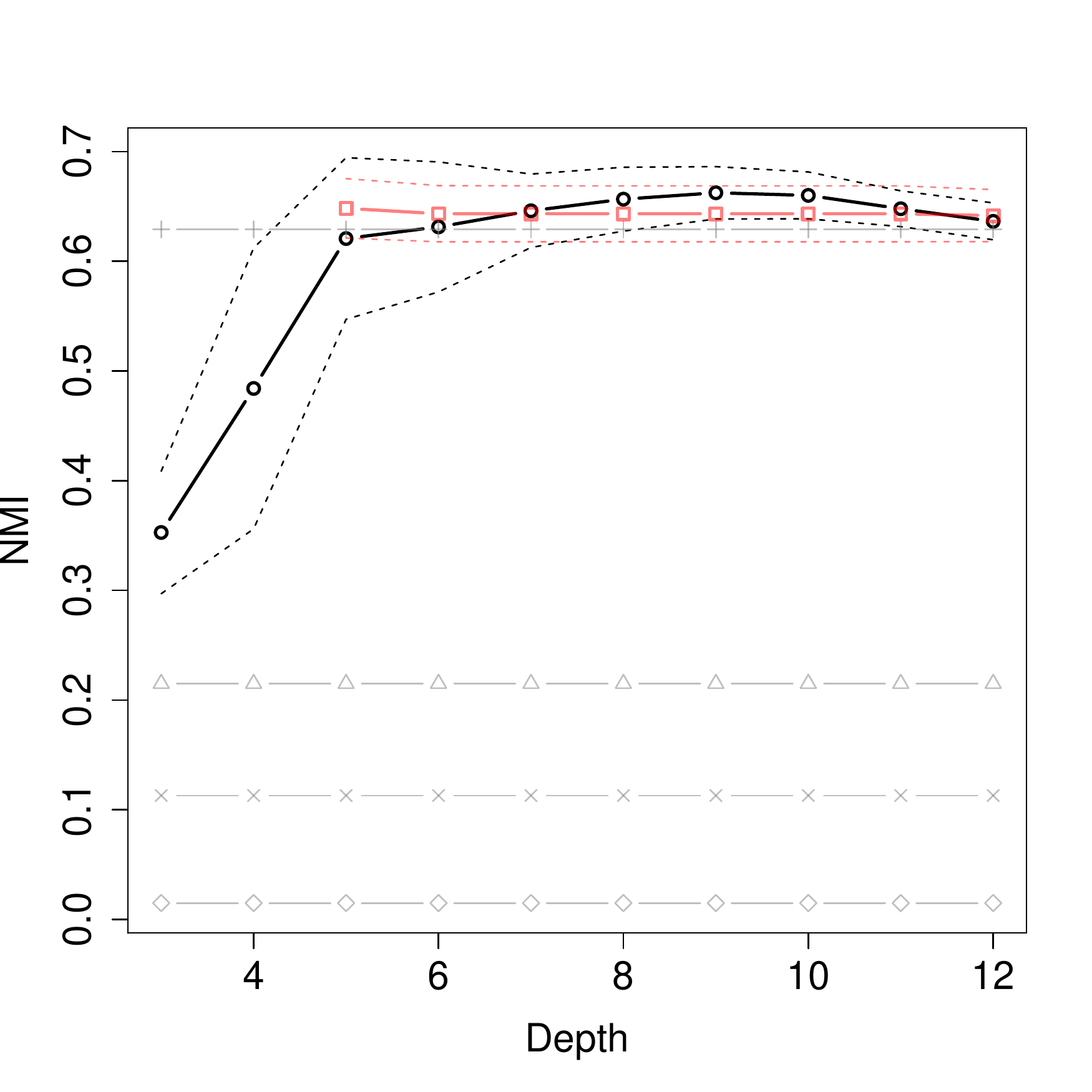}}
    \subfigure[Pendigits]{\includegraphics[width = 0.45\textwidth, height = 6cm]{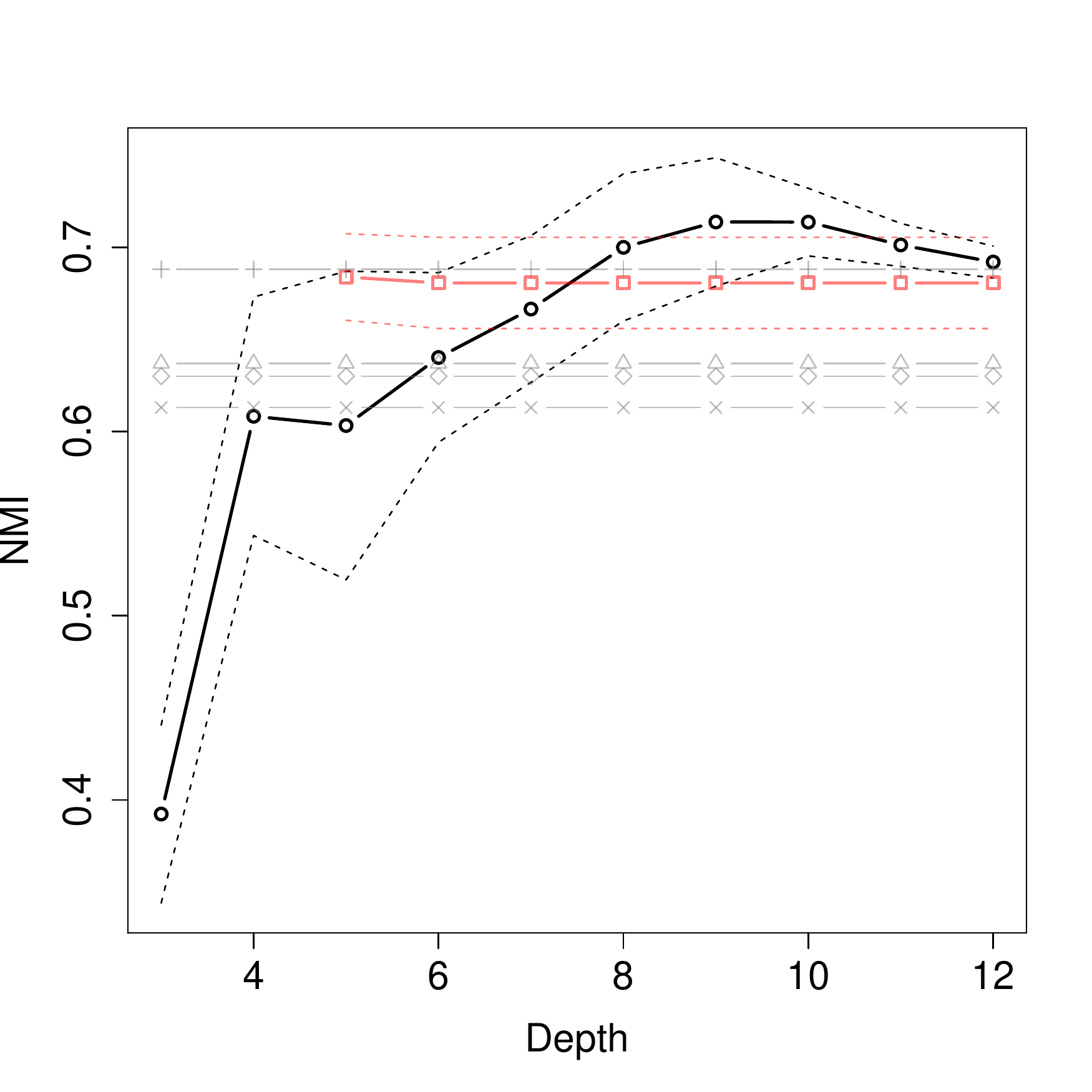}}
    \subfigure[Isolet]{\includegraphics[width = 0.45\textwidth, height = 6cm]{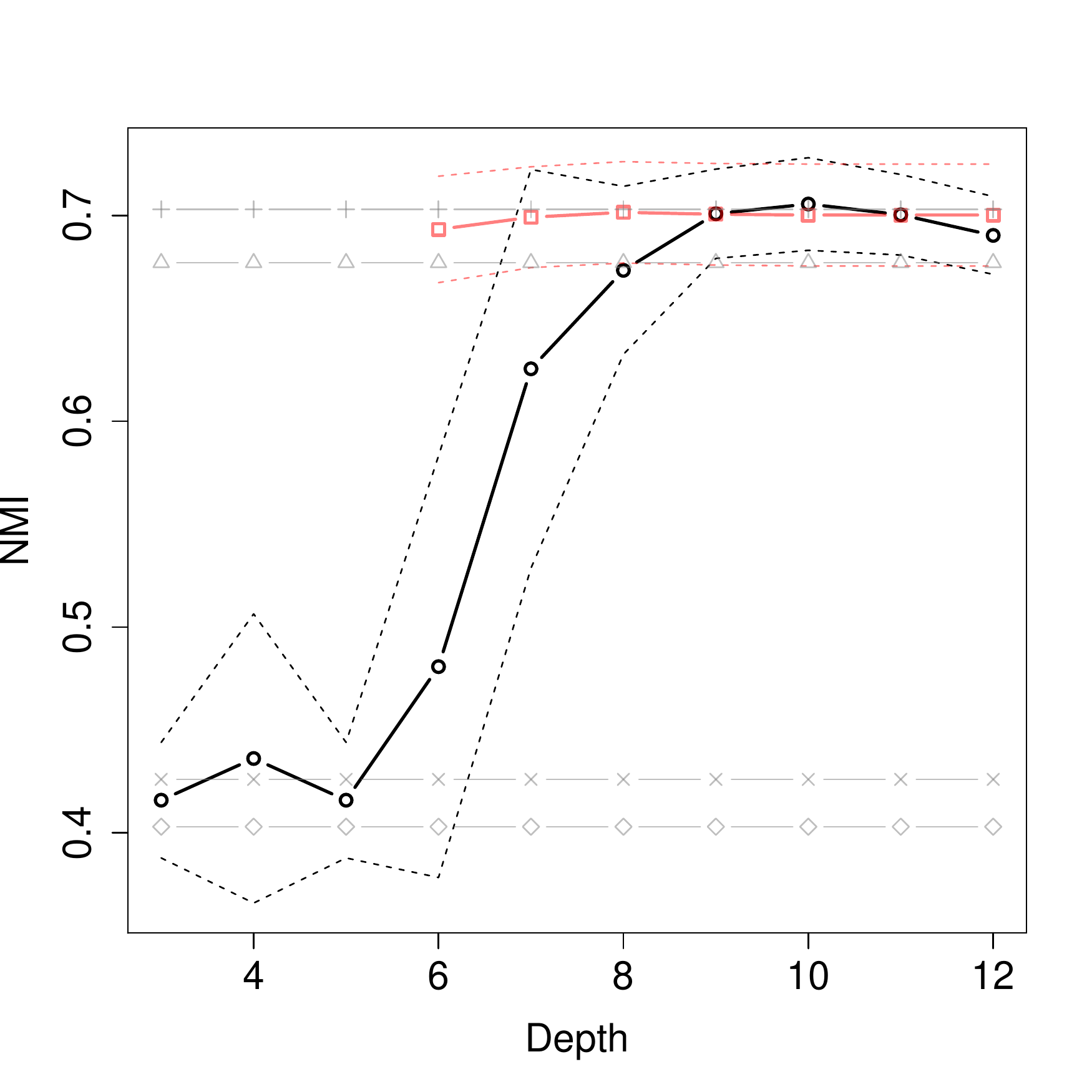}}
    \subfigure[Shuttle]{\includegraphics[width = 0.45\textwidth, height = 6cm]{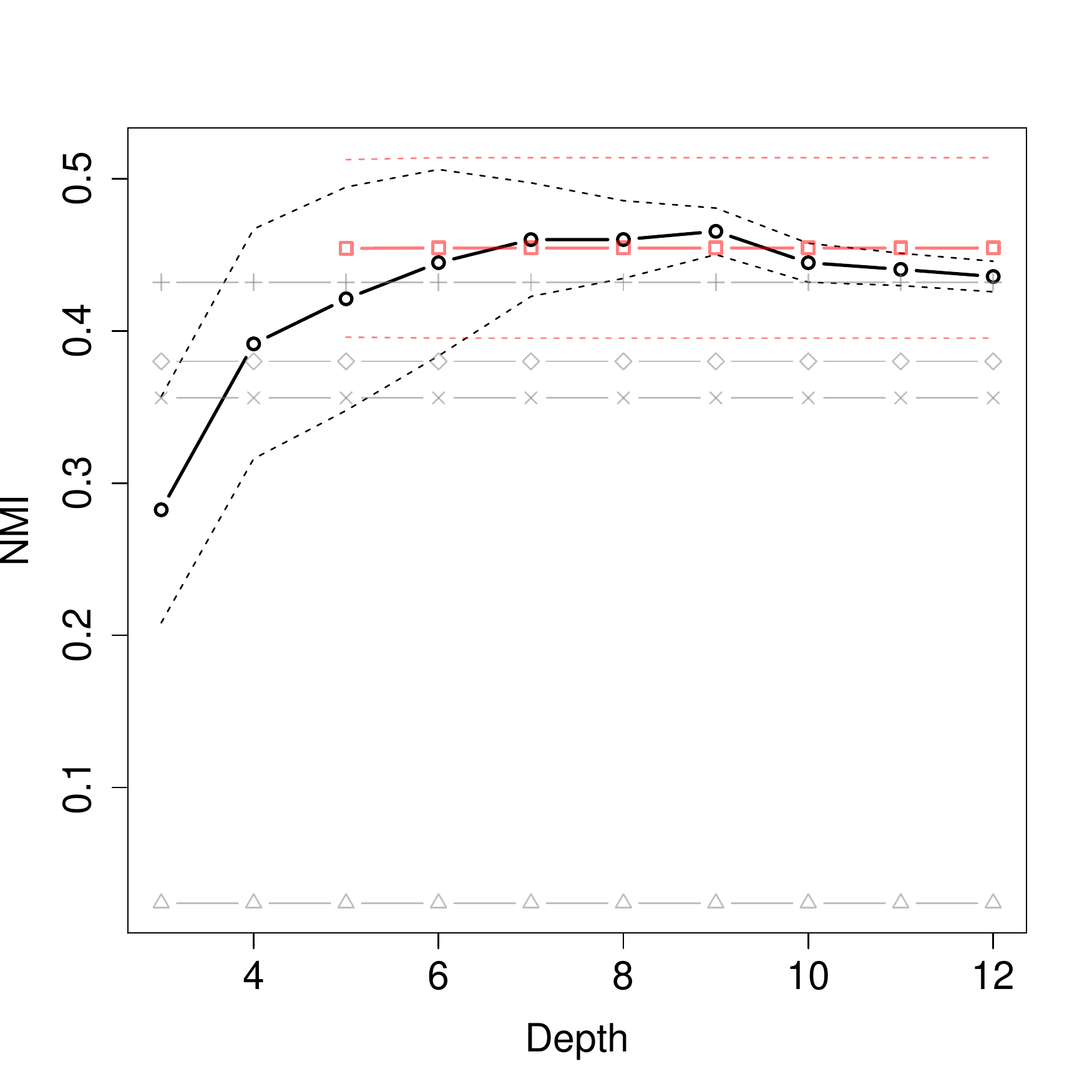}}
    \caption{Performance of iMDH (--$\circ$--) and iMDH$_k$ (\textcolor{red}{--{\small $\square$}--}) for varying tree depth based on NMI. In addition the performance of BICO (\textcolor{gray}{--$\triangle$--}), $k$-means (\textcolor{gray}{--$+$--}), SPDC (\textcolor{gray}{--$\diamond$--}) and dePDDP (\textcolor{gray}{--$\times$--}) are shown.}
    \label{fig:nmi_depth_sens}
\end{figure}

Figure~\ref{fig:nmi_depth_sens} shows plots of the average NMI scores (as well as one standard deviation bounds) for varying tree depth, based on 20 repetitions in which the randomness is induced by permuting the observations. The average performances from the other methods used for comparison are also indicated with the horizontal lines. In all cases trees of depth in the region of 7--9 tend to yield the best performance. However, it is worth noting that the range of numbers of ``true clusters'' in the data sets which we have considered does not include any very large numbers. We suspect that, when the number of clusters is 50 or more, an alternative approach similar to that adopted in SPDC, in which the tree is grown in an active and forward manner; adding additional nodes only when there is evidence of the need, may be preferable to the pruning approach we apply.

\begin{figure}[h]
    \centering
    \subfigure[Optidigits]{\includegraphics[width = 0.45\textwidth, height = 6cm]{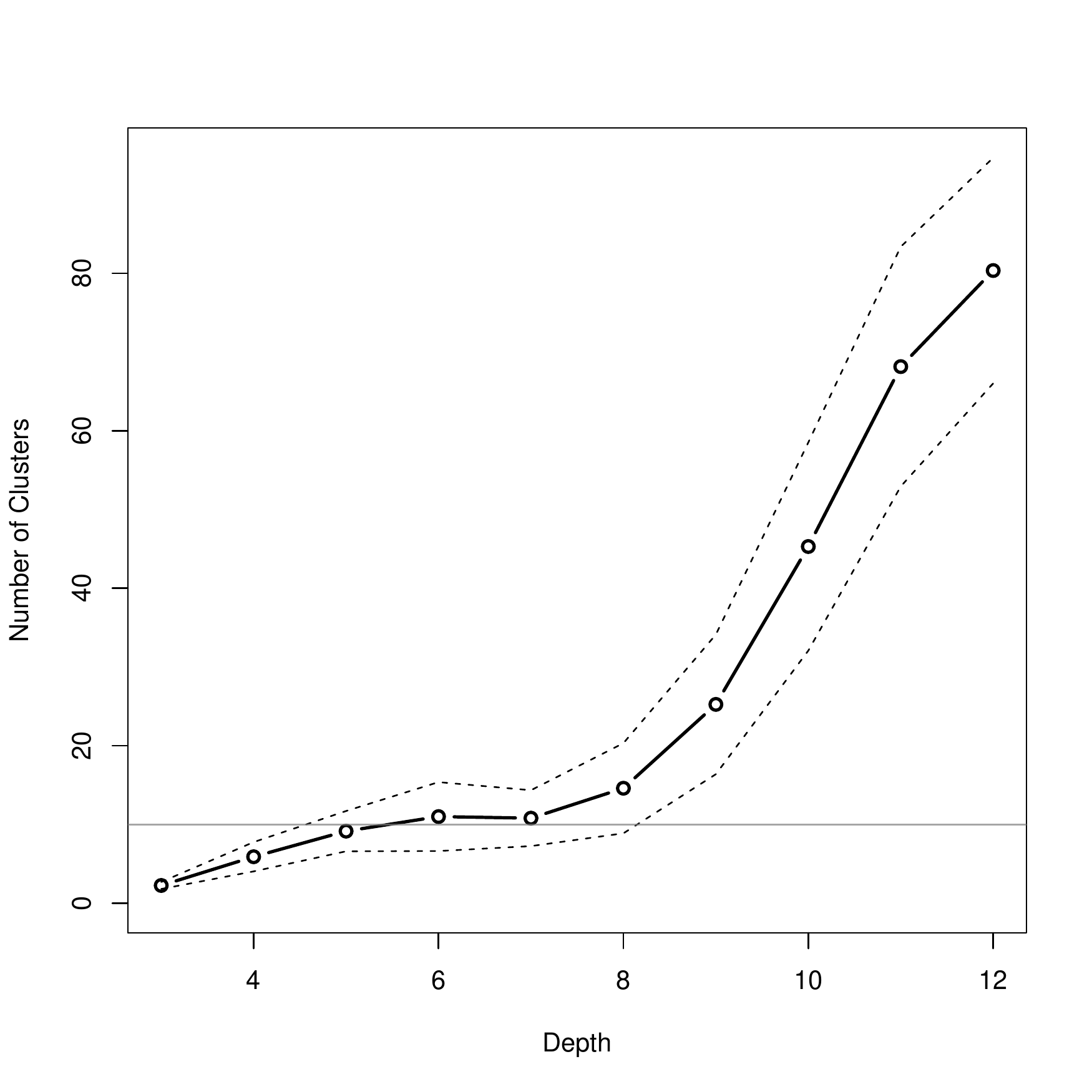}}
    \subfigure[Pendigits]{\includegraphics[width = 0.45\textwidth, height = 6cm]{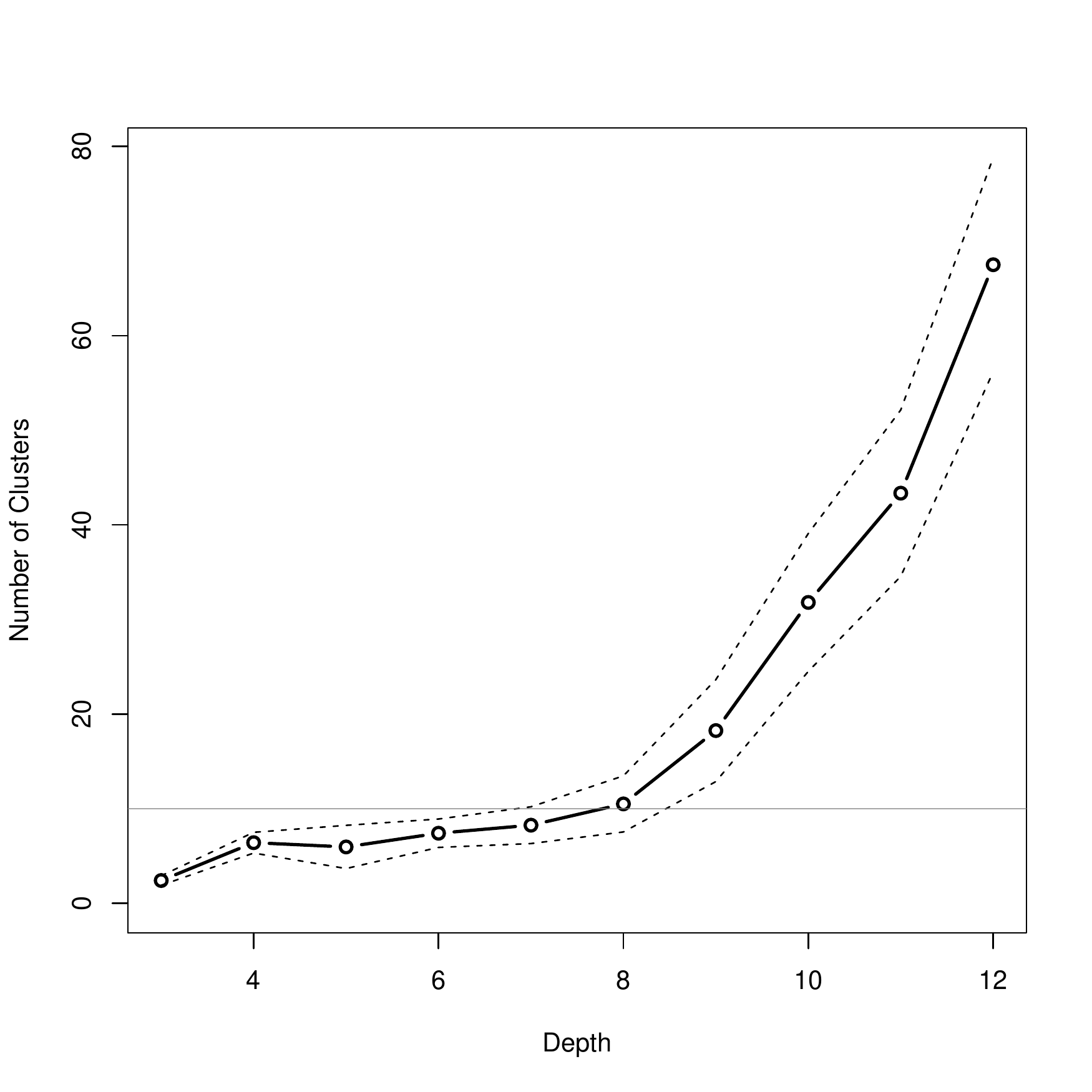}}
    \subfigure[Isolet]{\includegraphics[width = 0.45\textwidth, height = 6cm]{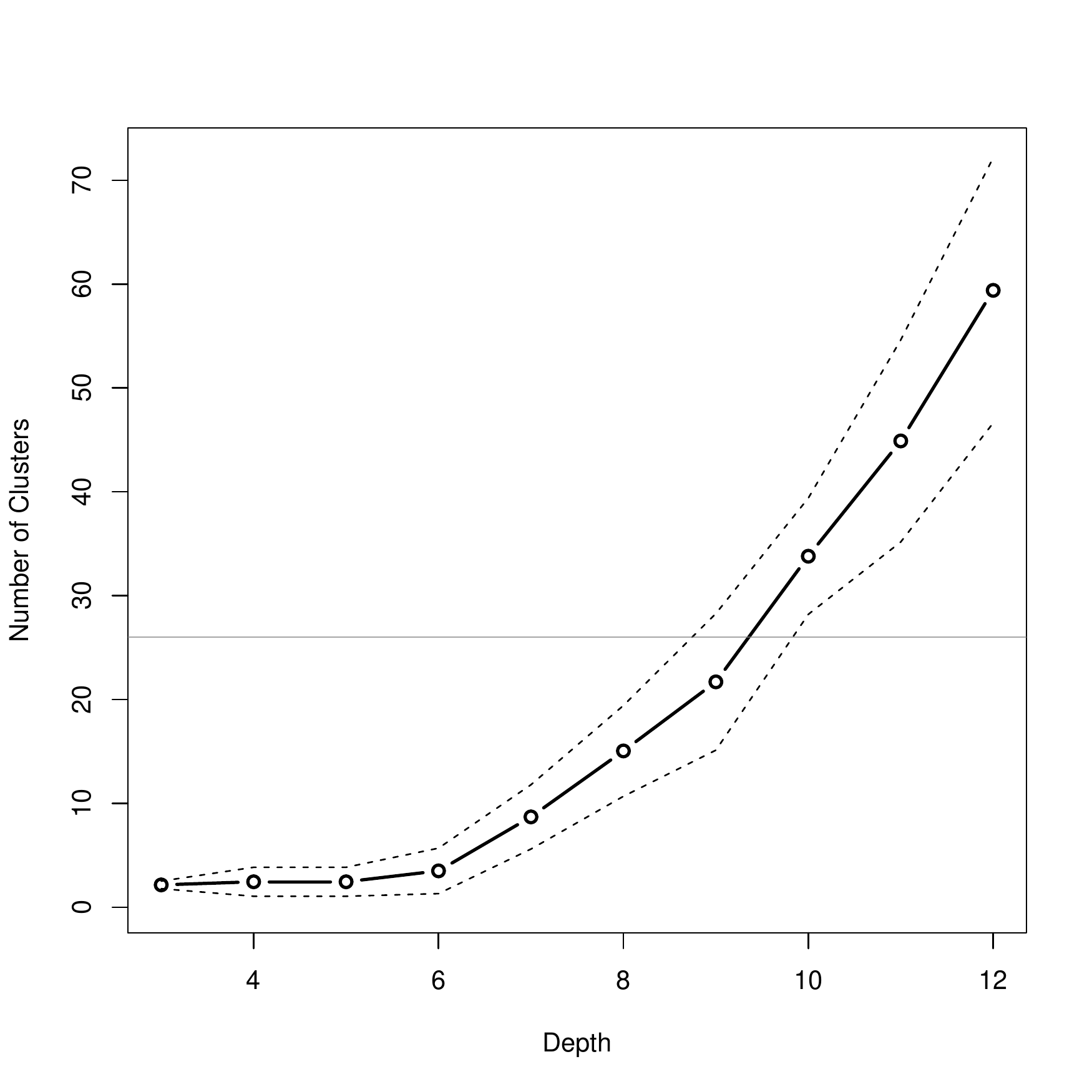}}
    \subfigure[Shuttle]{\includegraphics[width = 0.45\textwidth, height = 6cm]{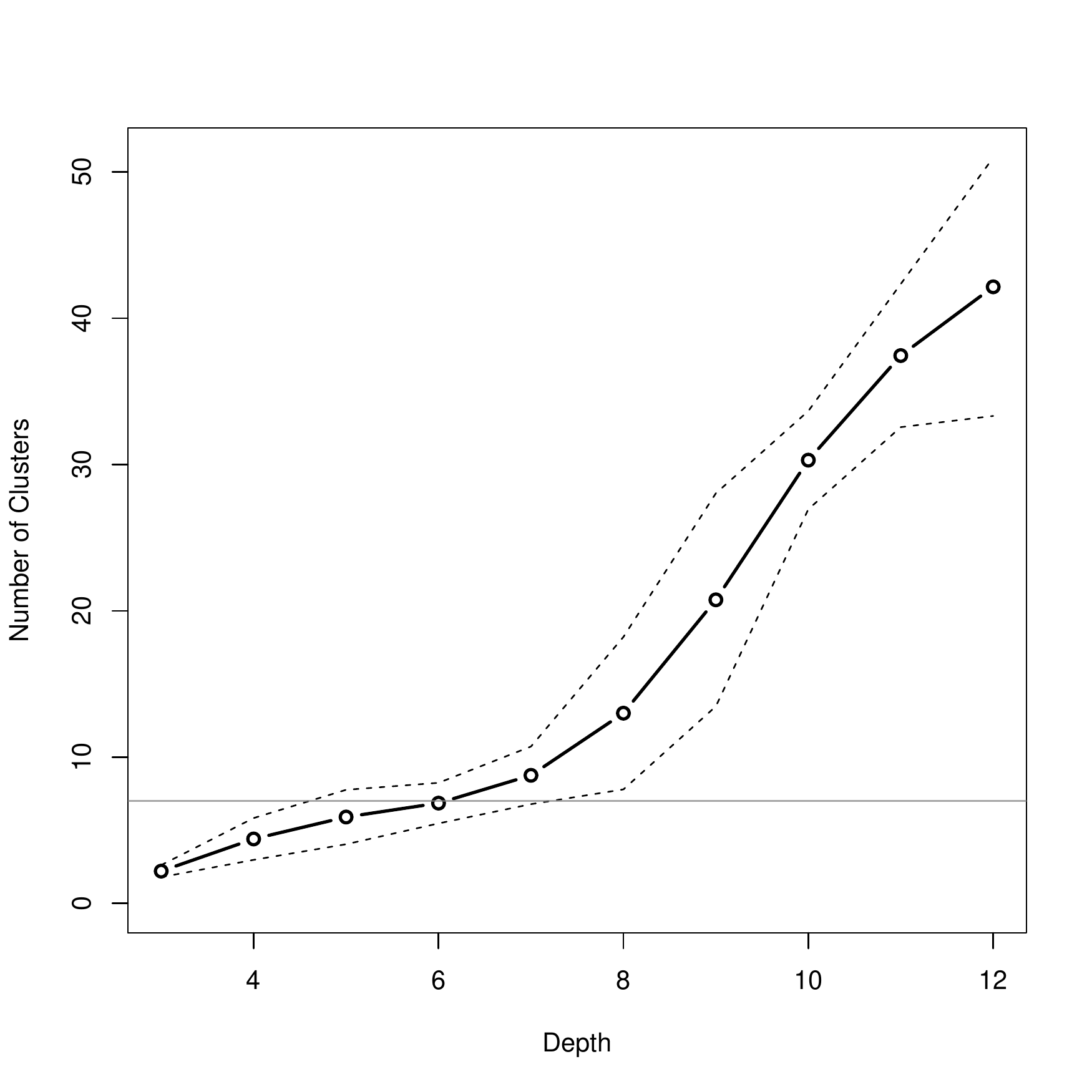}}
    \caption{Number of clusters selected by iMDH models for varying tree depth. Horizontal lines show the ``true'' number of clusters.}
    \label{fig:nc_depth_sens}
\end{figure}

Figure~\ref{fig:nc_depth_sens} shows plots of the average (with one standard error bounds) numbers of clusters selected from the same set of experiments, for varying tree depth.

\subsubsection{Varying Bandwidth Sequence}

\begin{figure}[h]
    \centering
    \subfigure[Optidigits]{\includegraphics[width = 0.45\textwidth, height = 6cm]{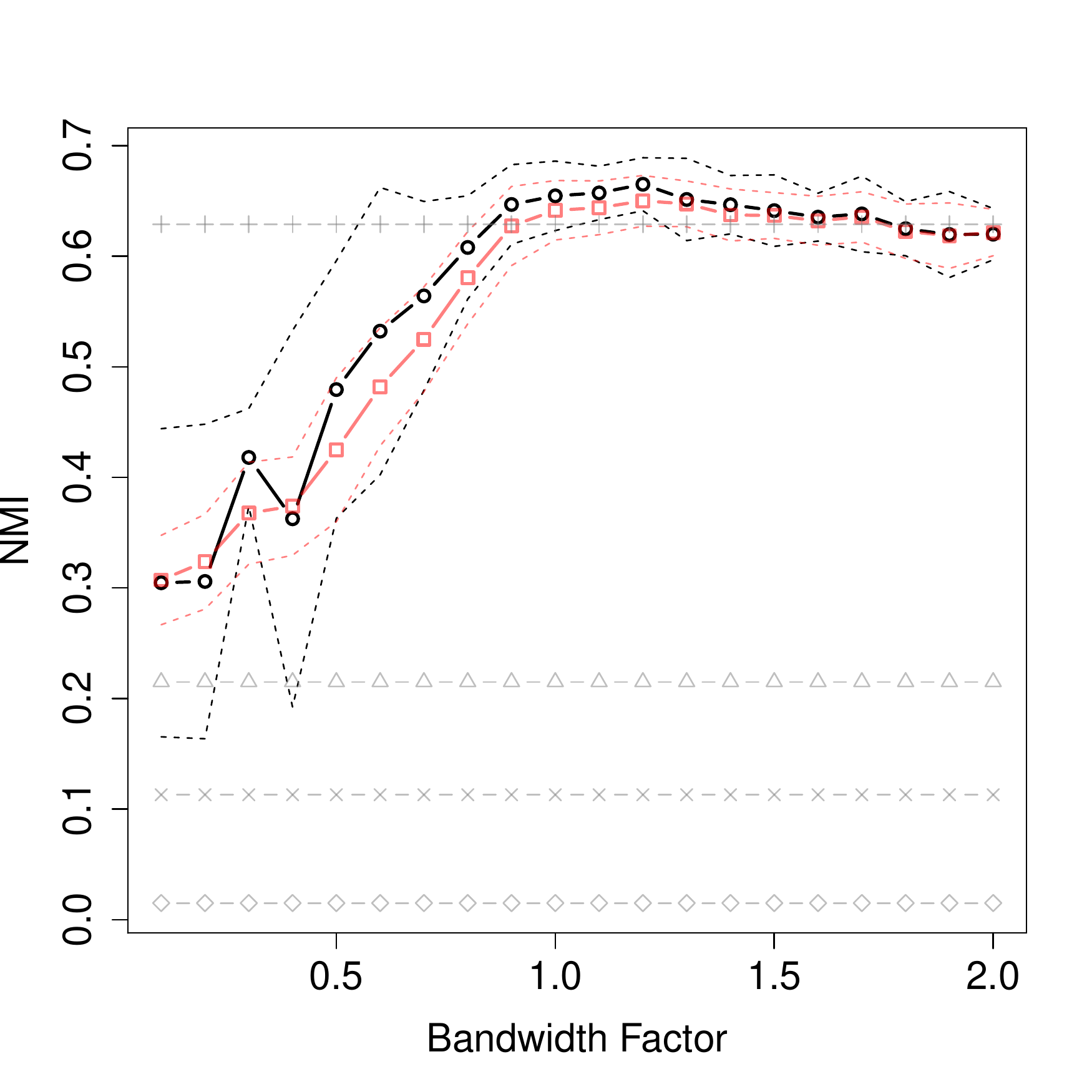}}
    \subfigure[Pendigits]{\includegraphics[width = 0.45\textwidth, height = 6cm]{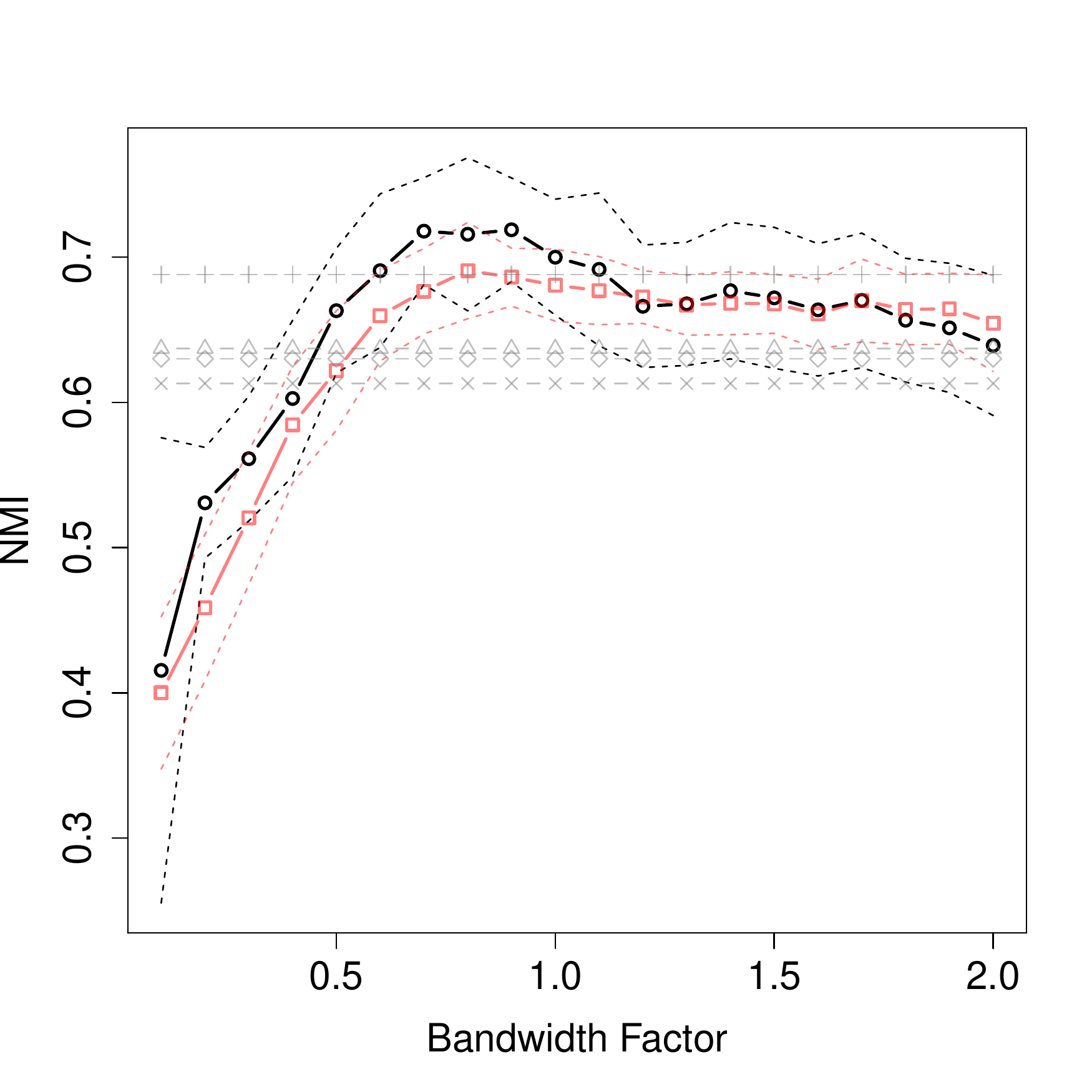}}
    \subfigure[Isolet]{\includegraphics[width = 0.45\textwidth, height = 6cm]{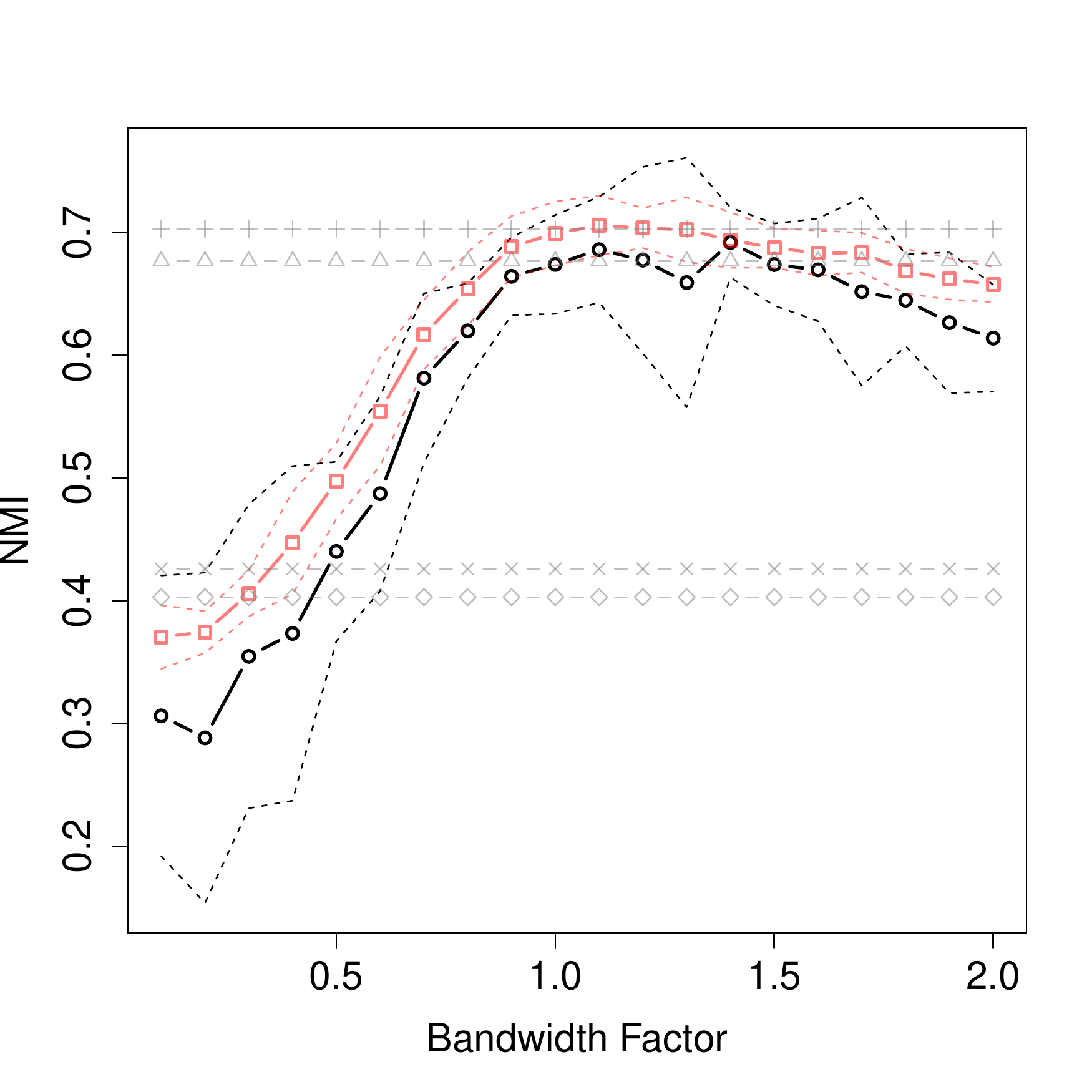}}
    \subfigure[Shuttle]{\includegraphics[width = 0.45\textwidth, height = 6cm]{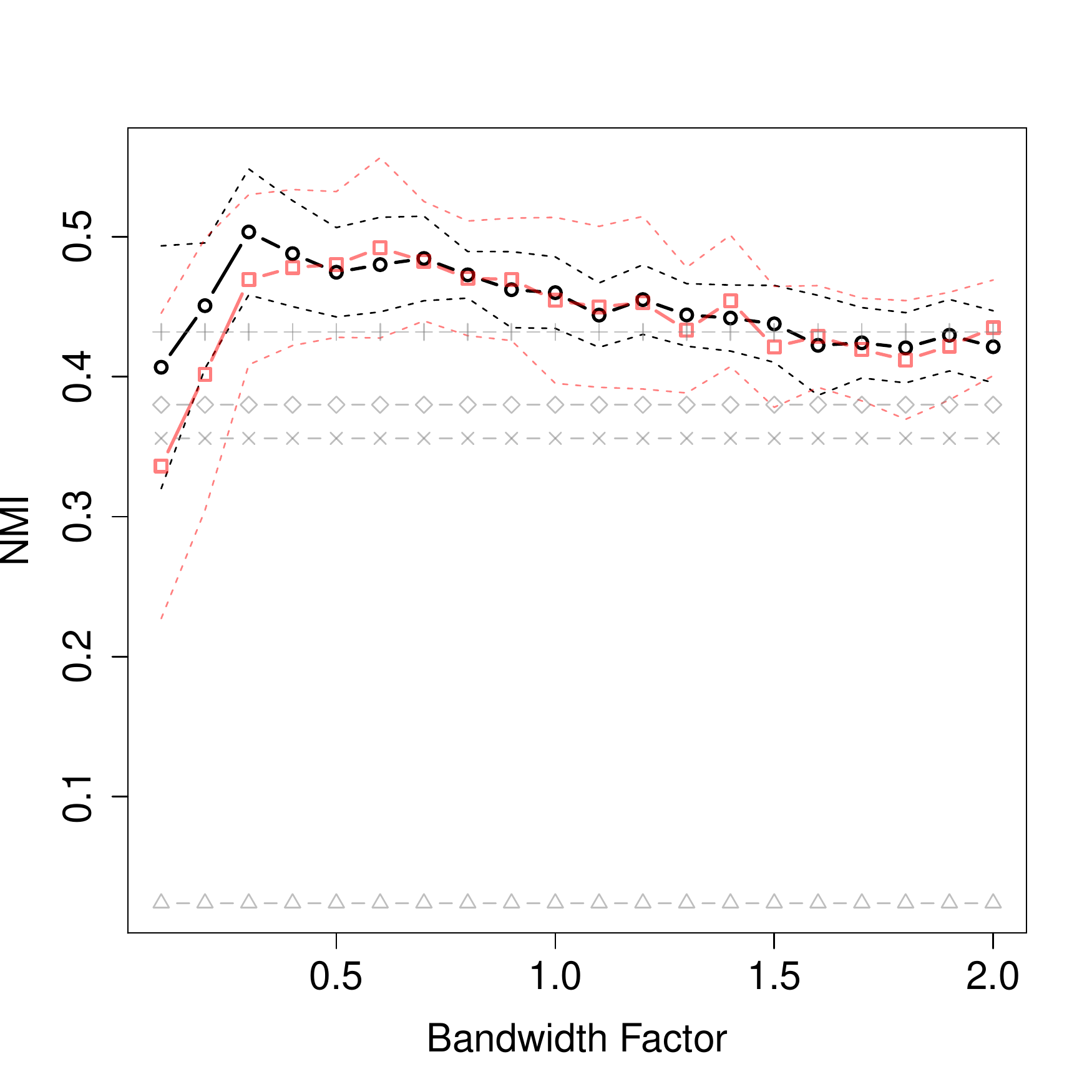}}
    \caption{Performance of iMDH (--$\circ$--) and iMDH$_k$ (\textcolor{red}{--{\small $\square$}--}) for varying bandwidth sequence based on NMI. In addition the performance of BICO (\textcolor{gray}{--$\triangle$--}), $k$-means (\textcolor{gray}{--$+$--}), SPDC (\textcolor{gray}{--$\diamond$--}) and dePDDP (\textcolor{gray}{--$\times$--}) are shown.}
    \label{fig:nmi_h_sens}
\end{figure}

\begin{figure}[h]
    \centering
    \subfigure[Optidigits]{\includegraphics[width = 0.45\textwidth, height = 6cm]{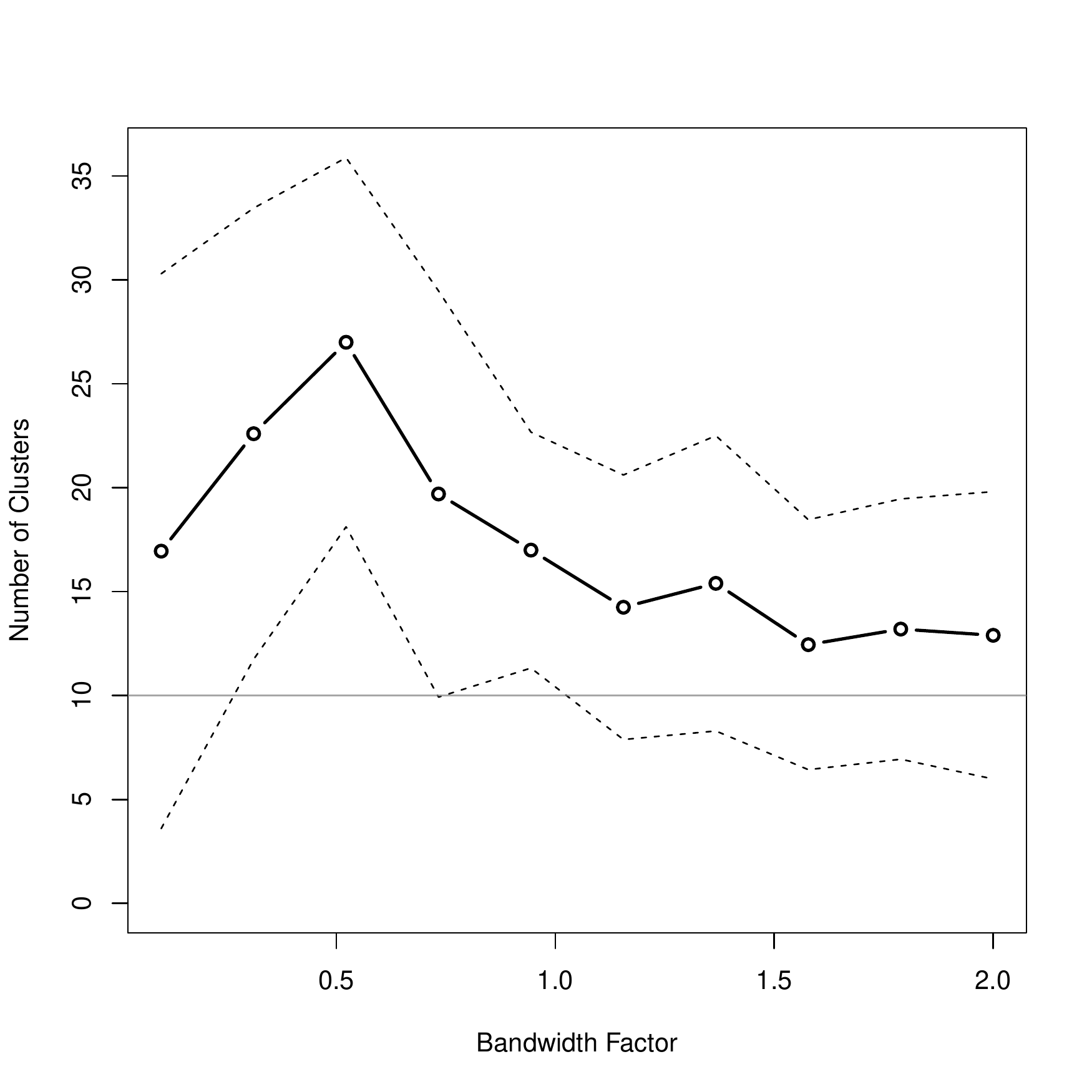}}
    \subfigure[Pendigits]{\includegraphics[width = 0.45\textwidth, height = 6cm]{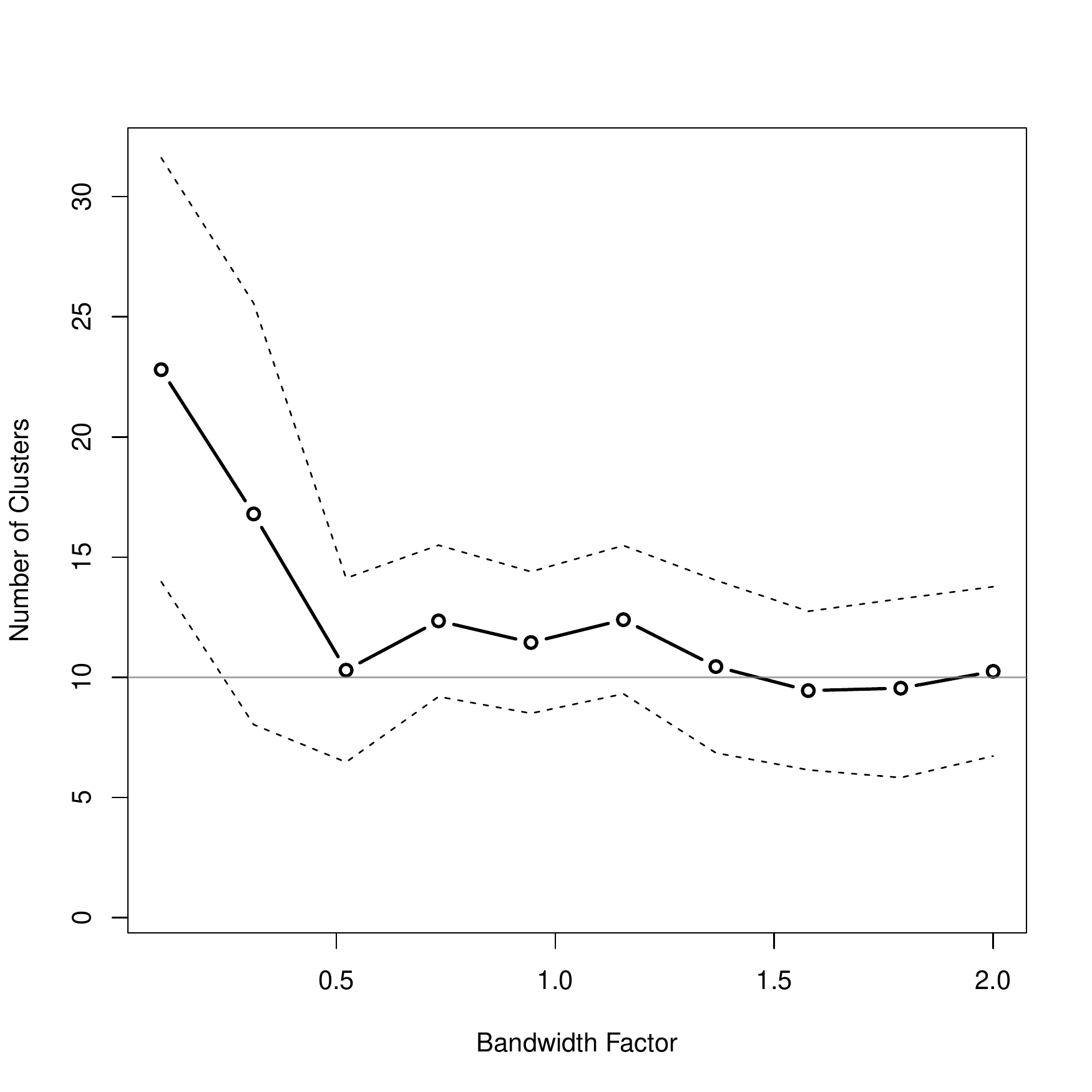}}
    \subfigure[Isolet]{\includegraphics[width = 0.45\textwidth, height = 6cm]{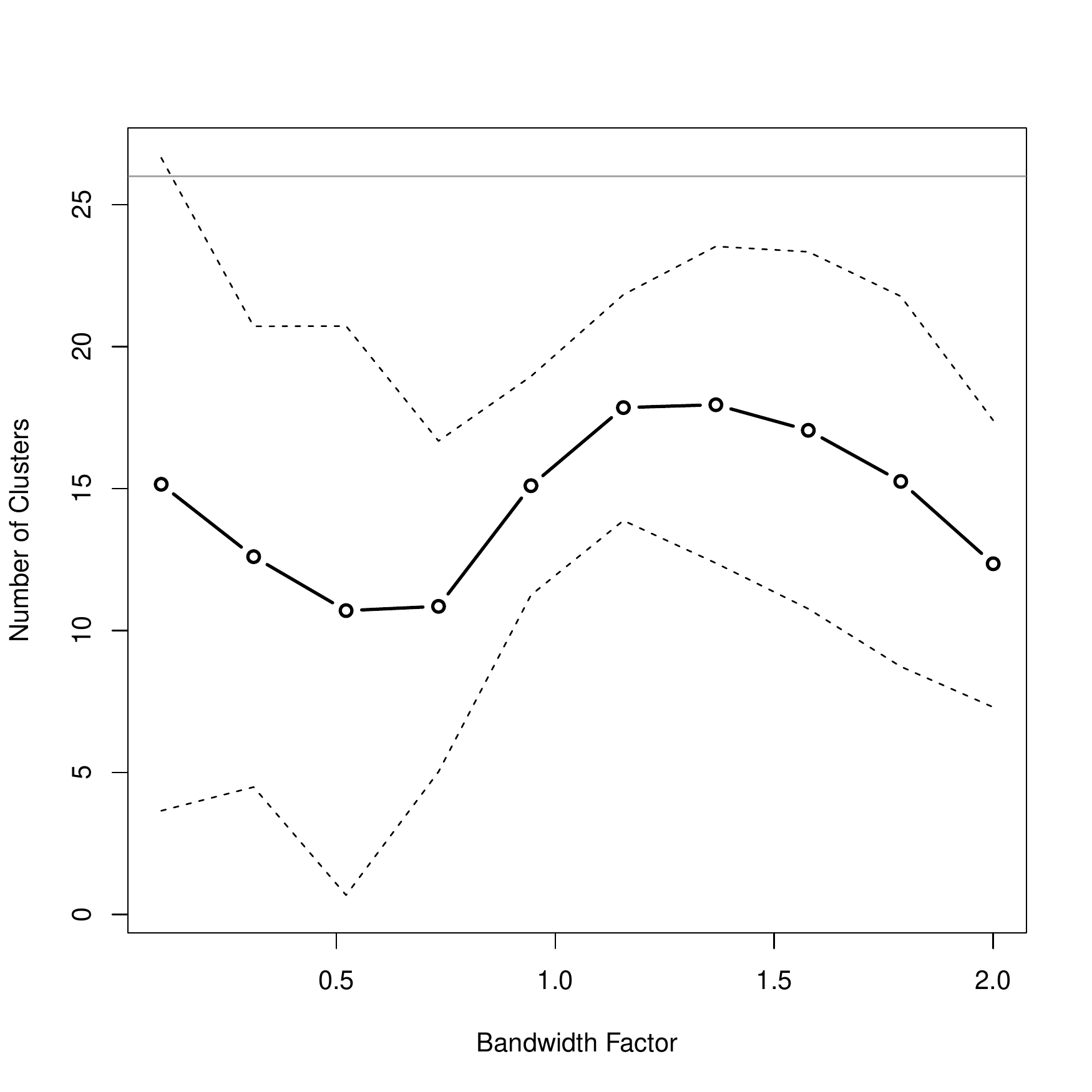}}
    \subfigure[Shuttle]{\includegraphics[width = 0.45\textwidth, height = 6cm]{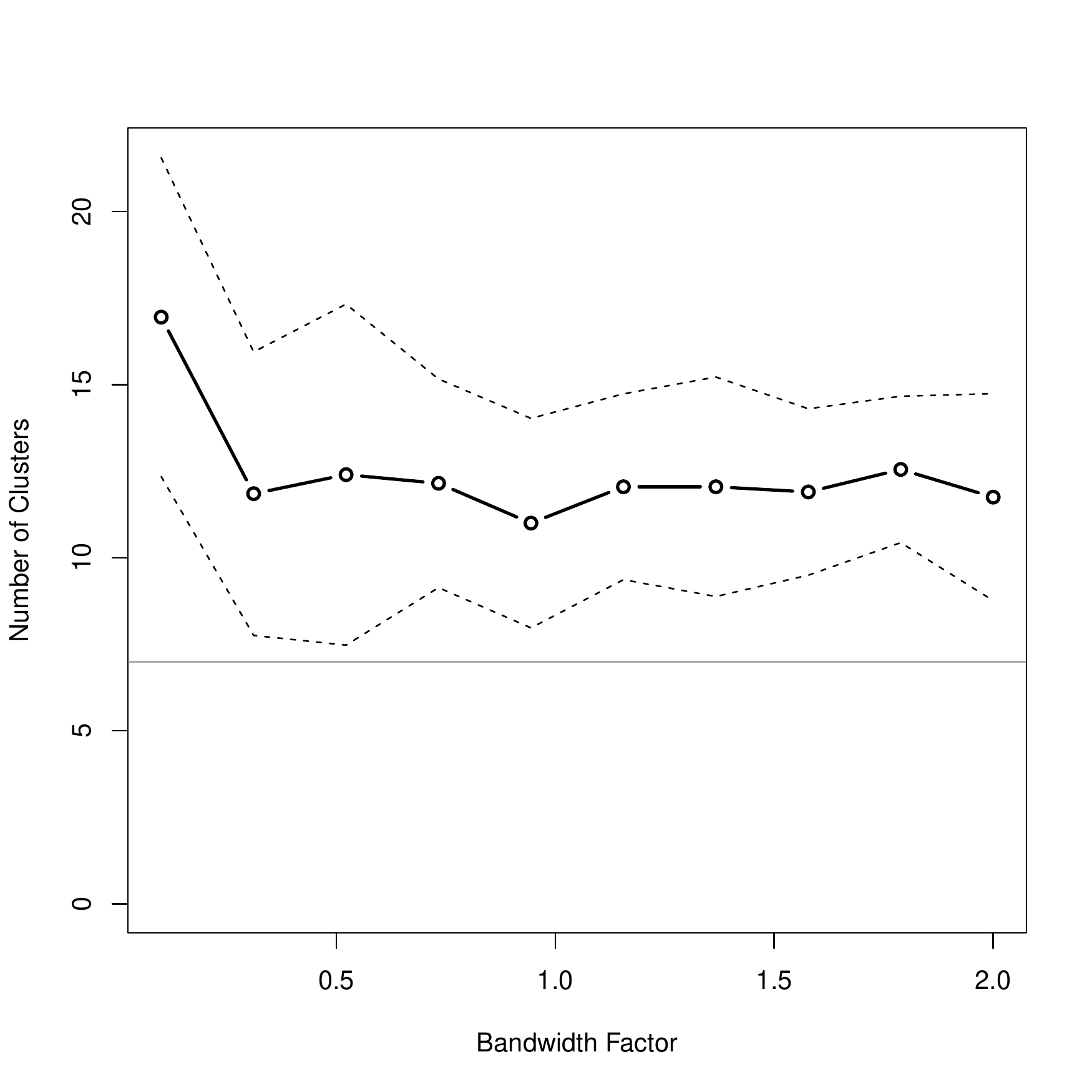}}
    \caption{Number of clusters selected by iMDH models for varying bandwidth sequence. Horizontal lines show the ``true'' number of clusters.}
    \label{fig:nc_h_sens}
\end{figure}

It is intuitively the case that setting the bandwidth sequence along the lines of what is used in univariate kernel density estimation (KDE) is sensible, since the smoothing used in the proposed method is conducted along univariate projections. Furthermore, we found that the optimal rate of $t^{-0.2}$ for univariate KDE is consistent with convergence of the method. The value $h^{(t)} = \hat \sigma_{\v^{(t)}}t^{-0.2}$ therefore is intuitively reasonable and consistent with our theory. Here we explore variations to this by setting $h^{(t)} = \delta\hat \sigma_{\v^{(t)}}t^{-0.2}$ for a range of factors $\delta \in [0.1, 2]$. For simplicty's sake we use the same four data sets as those used in the previous subsection. The results can be seen in Figures~\ref{fig:nmi_h_sens} and~\ref{fig:nc_h_sens}.
As mentioned previously, smaller values of the bandwidth will generally lead to faster but more variable learning. This is somewhat reflected in the results, where the number of clusters selected by the method is fairly stable across different values for $\delta$ (Figure~\ref{fig:nc_h_sens}), but the performance is markedly worse for very small values (Figure~\ref{fig:nmi_h_sens}). This suggests inaccurate estimation of cluster boundaries, rather than inappropriate selection of the number of clusters, which is leading to poor performance.

\subsubsection{Varying $\alpha$}

Recall that $\alpha$ constrains the distance from the hyperplane to the mean of the observations, and hence smaller values for $\alpha$ tend to lead to more balanced splits in the hierarchical model, but also have the potential to restrict the learning so that cluster boundaries do not fall into regions of minimum density. A consequence of this is that even when high density regions in the underlying distribution are linearly separable it is possible the parts of these high density regions are ``cut off'' from their majority if $\alpha$ is set too small. While the hierarchical model structure means that these parts can subsequently be separated from other clusters, selection of $\alpha$ may best be performed along with any available domain knowledge. For example, if partitioning a single high density region into multiple clusters is acceptable, then this behaviour based on small $\alpha$ is very acceptable. On the other hand, if ``true clusters'' should be kept complete in spite of the possible risk that multiple ``true clusters'' are merged in the solution obtained, then setting $\alpha$ larger may be preferable.

\begin{figure}[h]
    \centering
    \subfigure[Optidigits]{\includegraphics[width = 0.45\textwidth, height = 6cm]{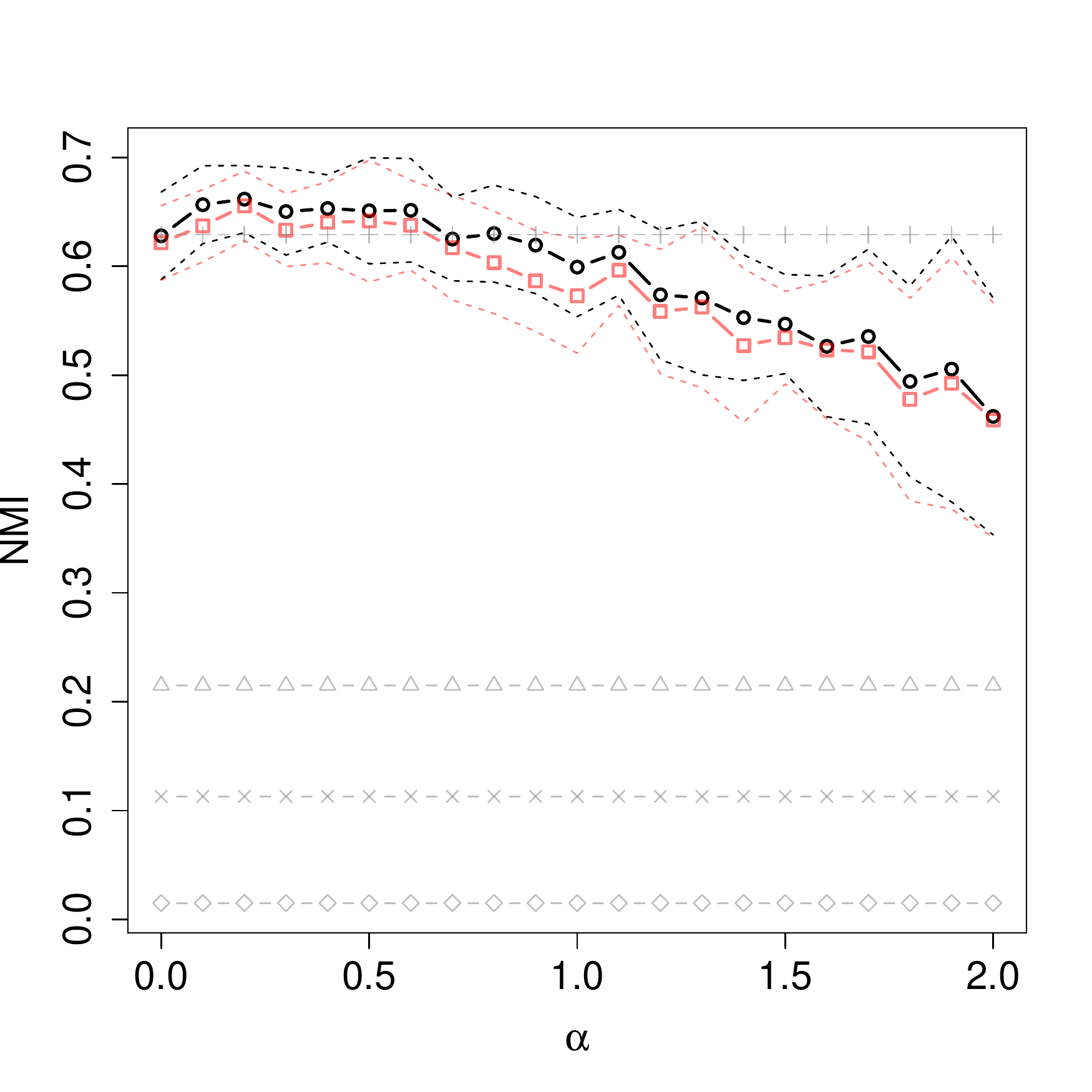}}
    \subfigure[Pendigits]{\includegraphics[width = 0.45\textwidth, height = 6cm]{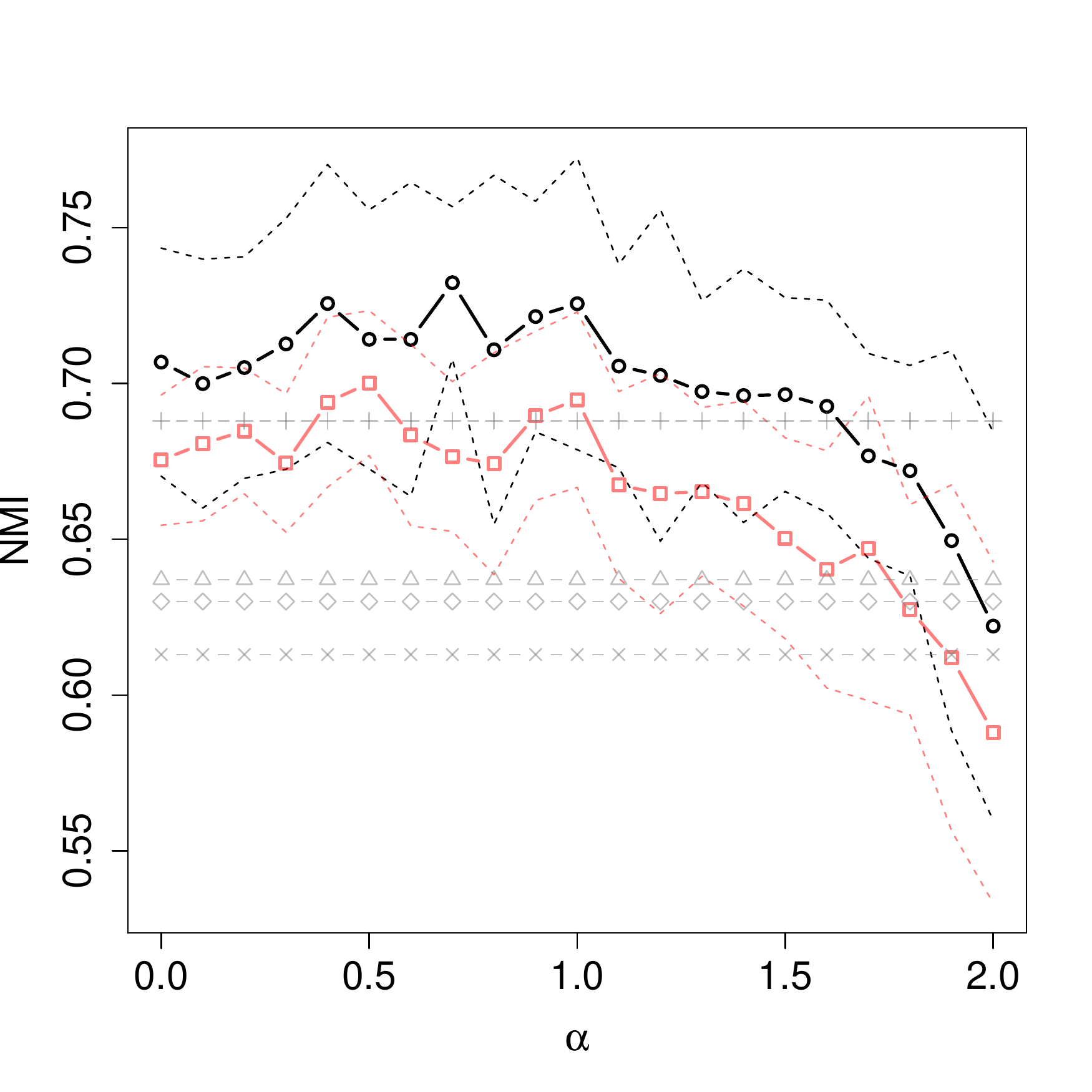}}
    \subfigure[Isolet]{\includegraphics[width = 0.45\textwidth, height = 6cm]{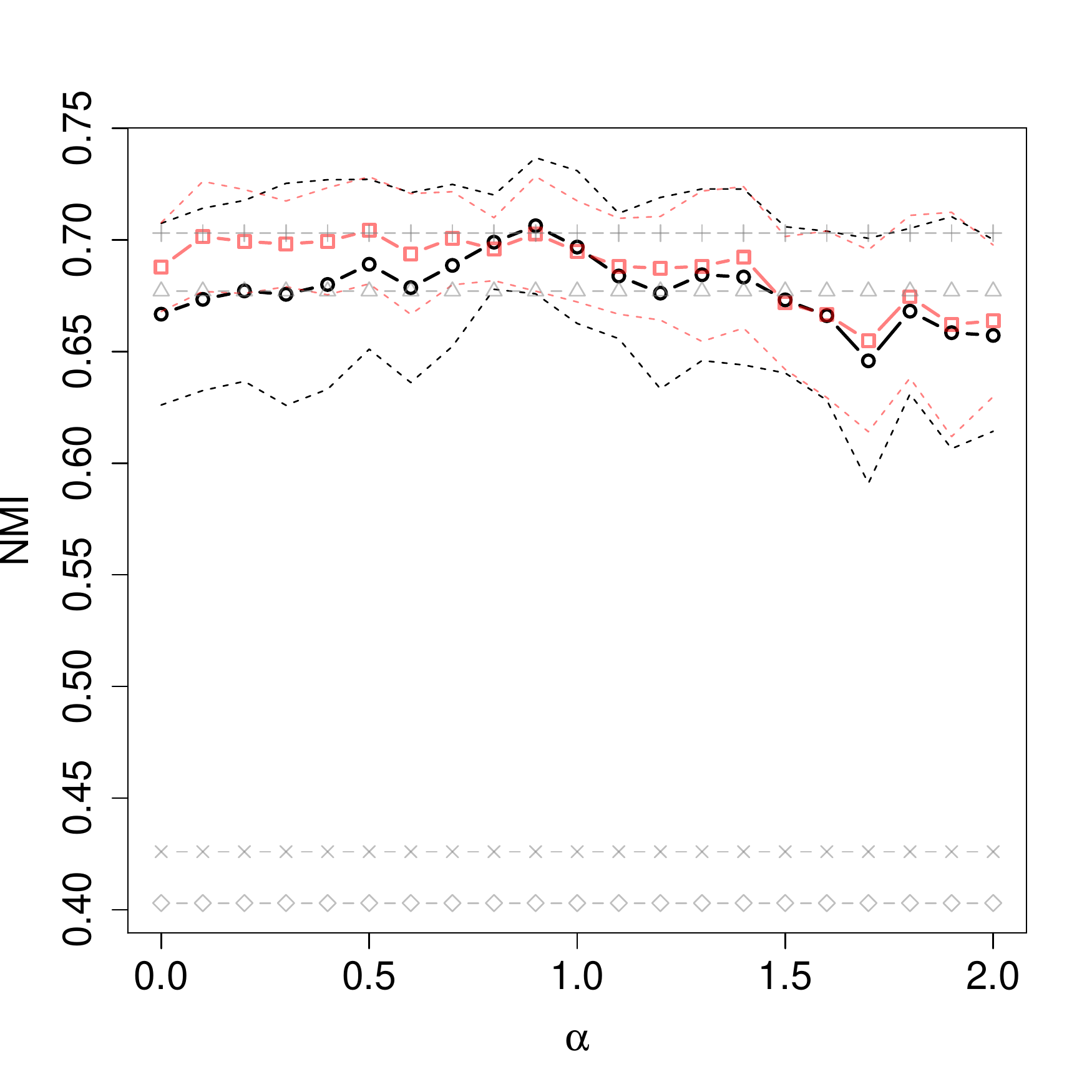}}
    \subfigure[Shuttle]{\includegraphics[width = 0.45\textwidth, height = 6cm]{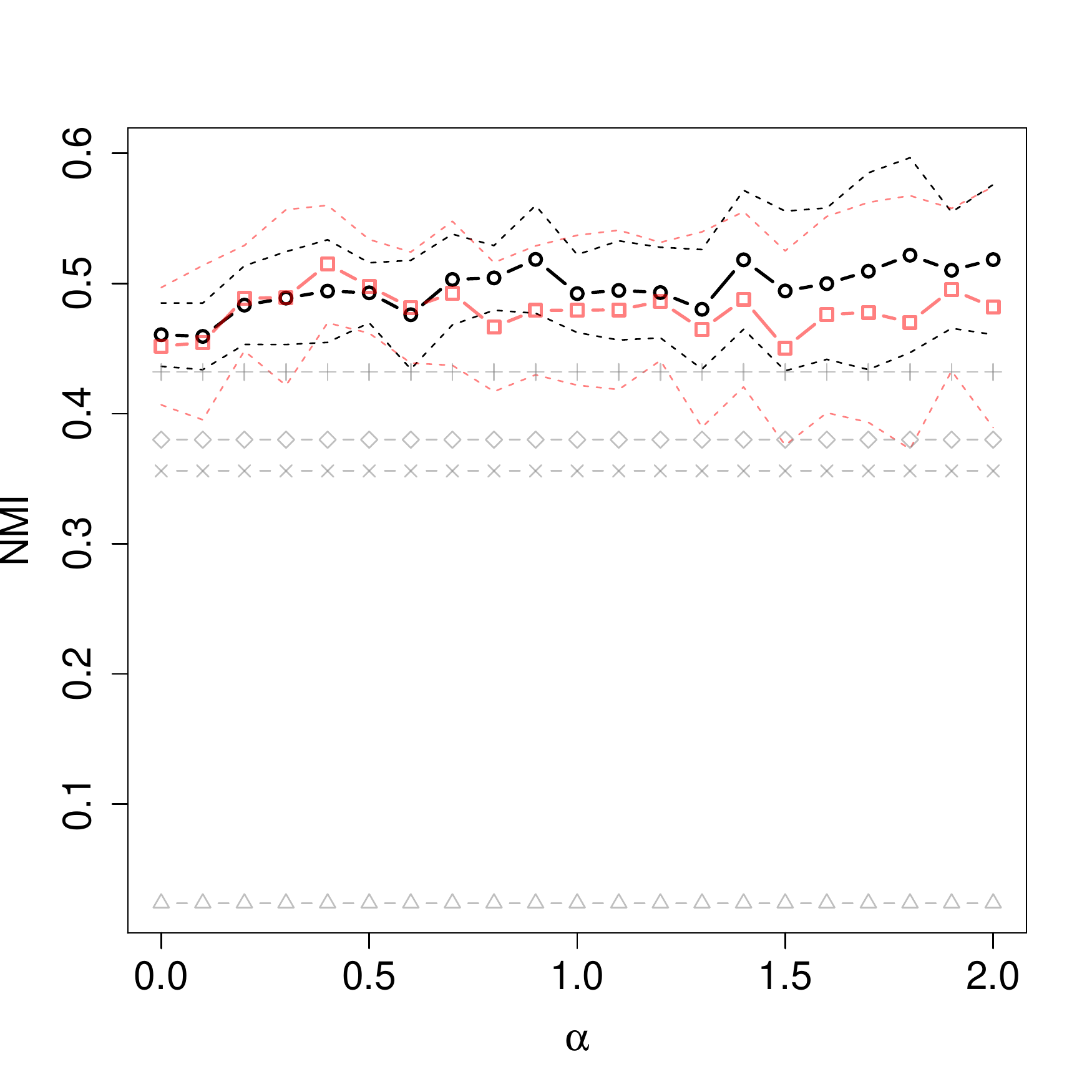}}
    \caption{Performance of iMDH (--$\circ$--) and iMDH$_k$ (\textcolor{red}{--{\small $\square$}--}) for varying $\alpha$ based on NMI. In addition the performance of BICO (\textcolor{gray}{--$\triangle$--}), $k$-means (\textcolor{gray}{--$+$--}), SPDC (\textcolor{gray}{--$\diamond$--}) and dePDDP (\textcolor{gray}{--$\times$--}) are shown.}
    \label{fig:nmi_alpha_sens}
\end{figure}

\begin{figure}[h]
    \centering
    \subfigure[Optidigits]{\includegraphics[width = 0.45\textwidth, height = 6cm]{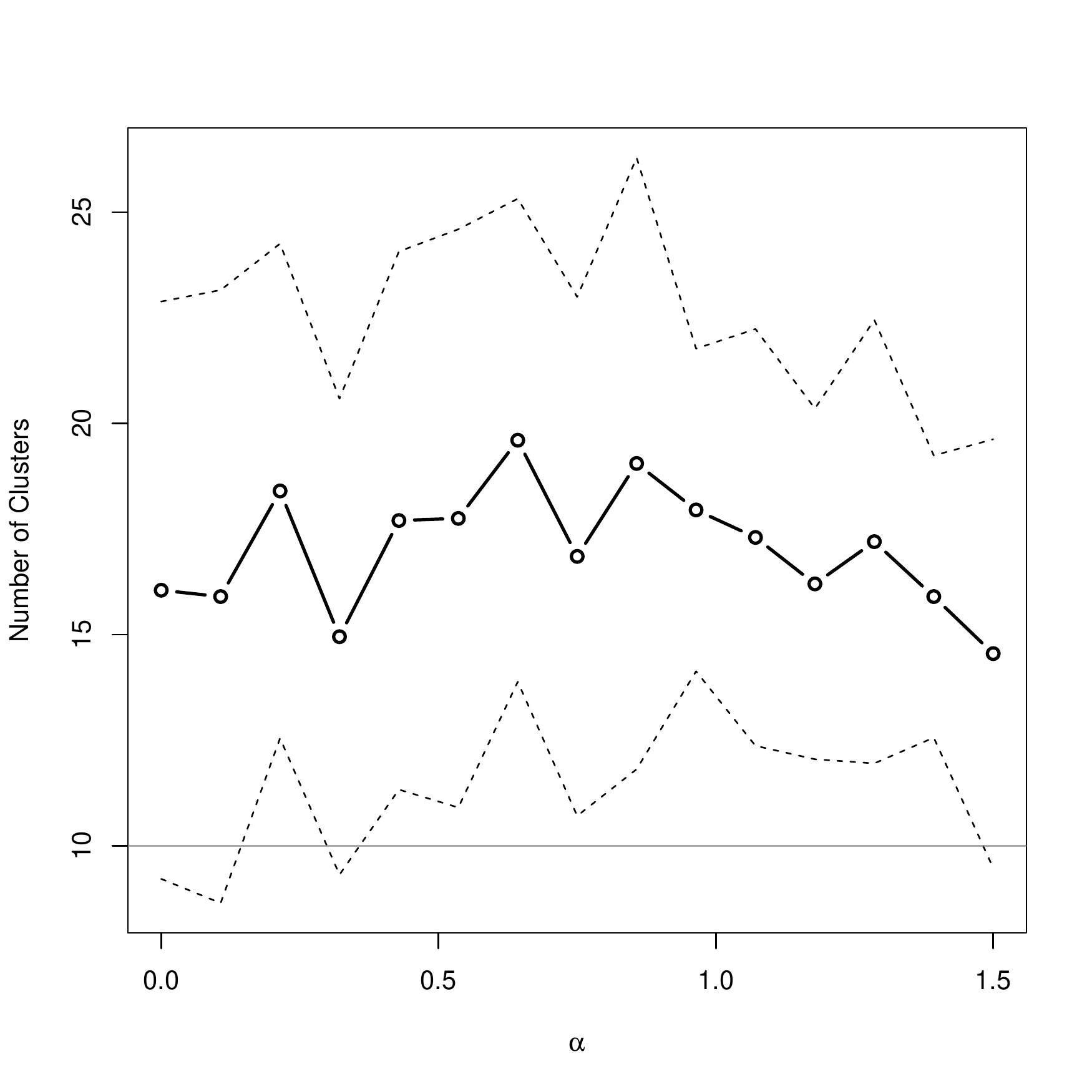}}
    \subfigure[Pendigits]{\includegraphics[width = 0.45\textwidth, height = 6cm]{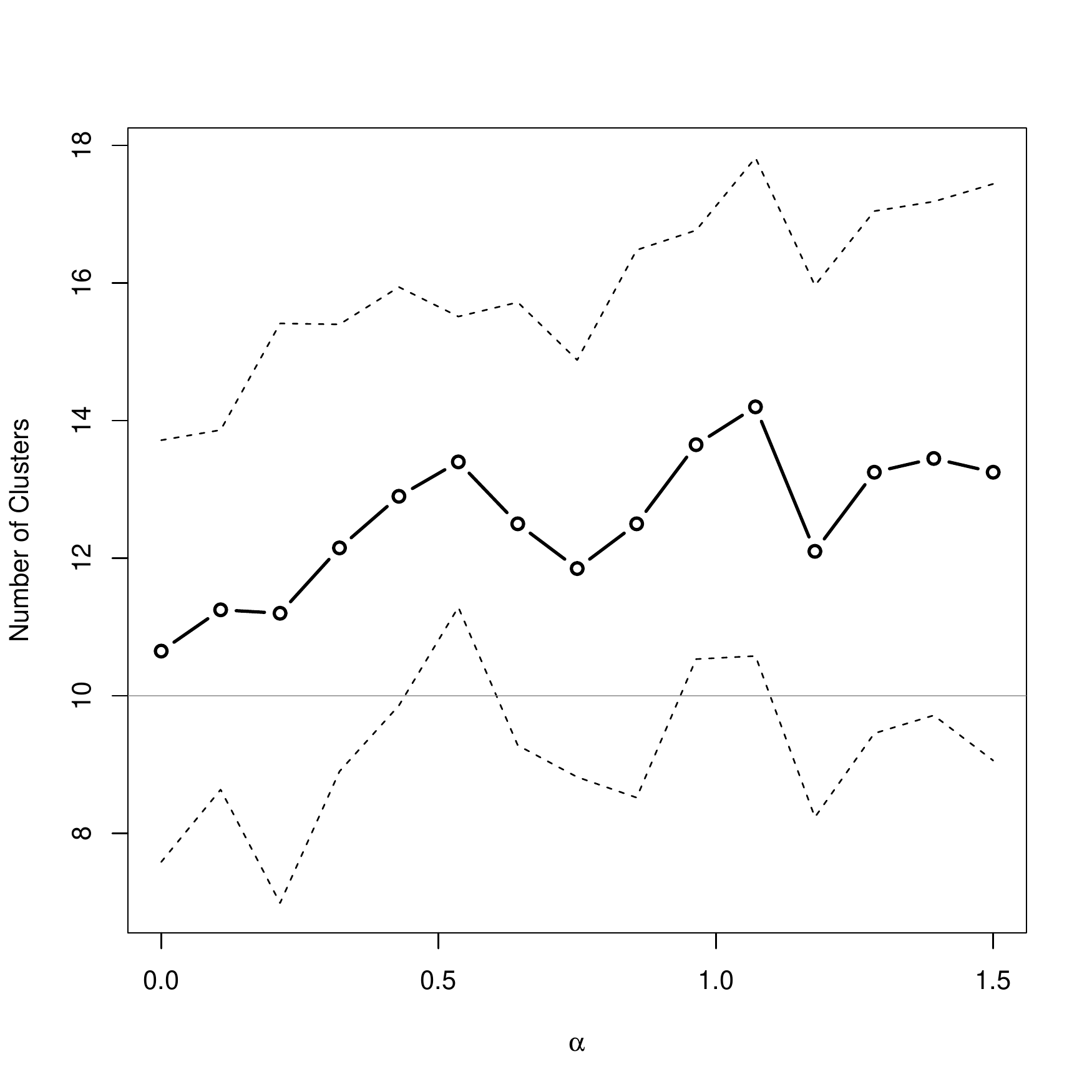}}
    \subfigure[Isolet]{\includegraphics[width = 0.45\textwidth, height = 6cm]{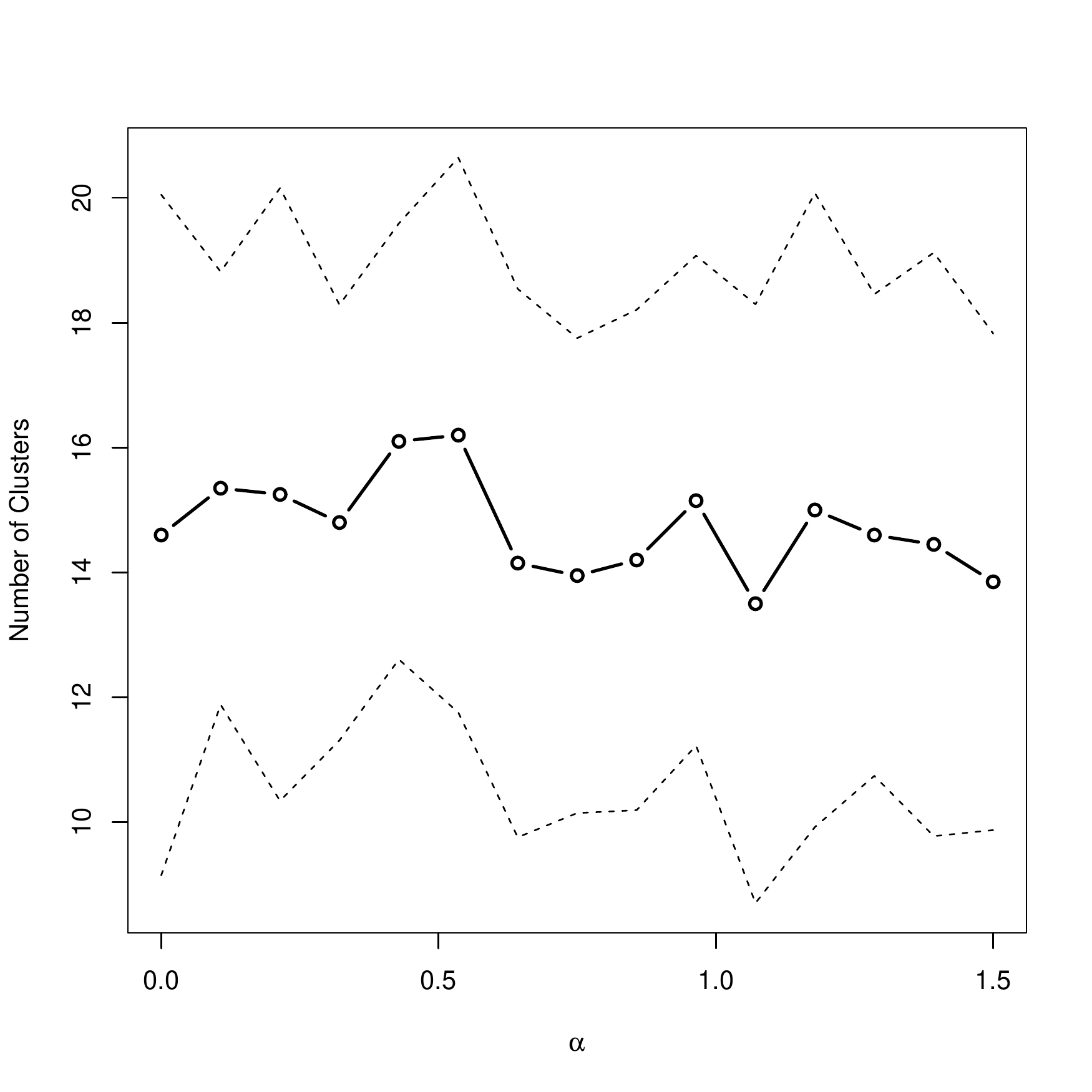}}
    \subfigure[Shuttle]{\includegraphics[width = 0.45\textwidth, height = 6cm]{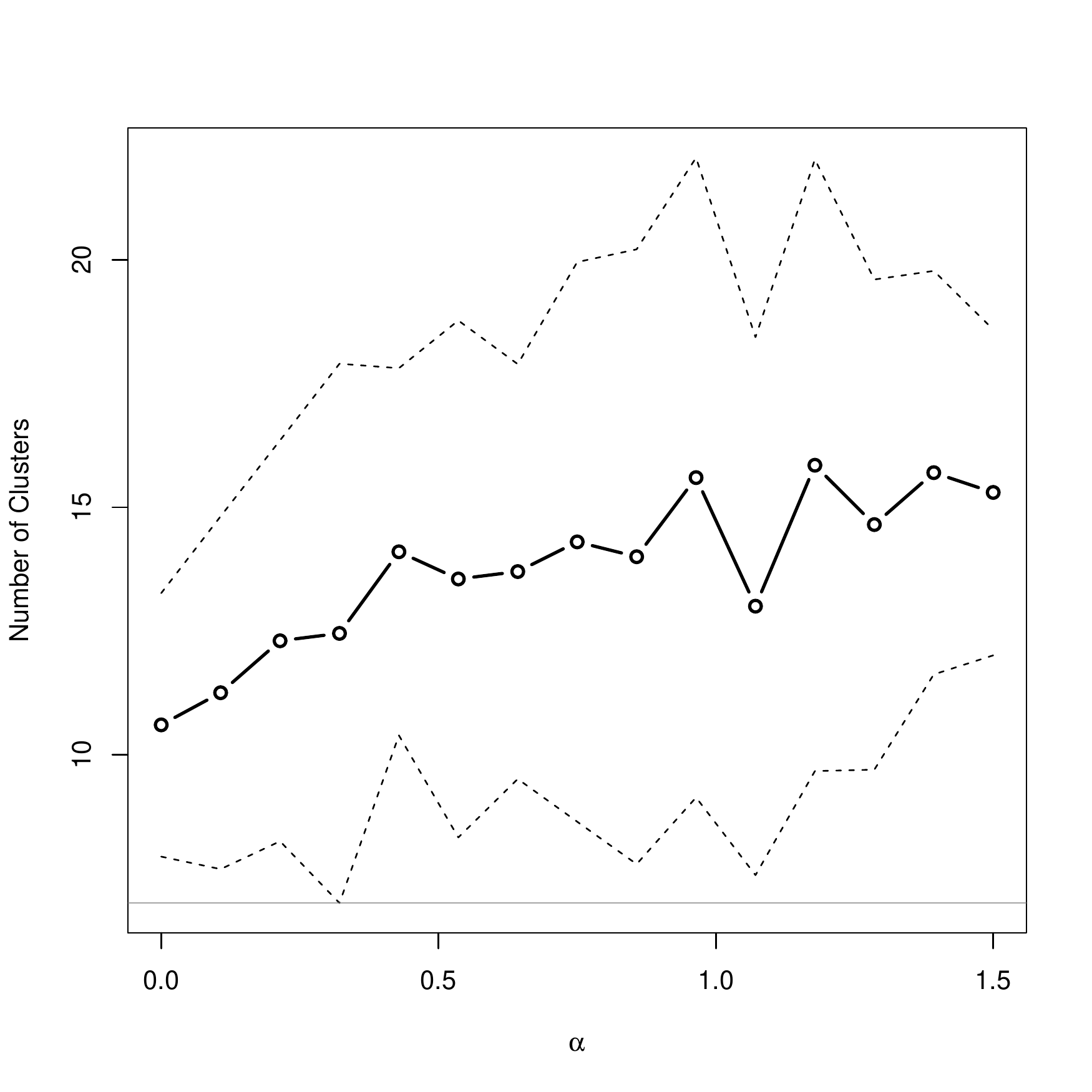}}
    \caption{Number of clusters selected by iMDH models for varying $\alpha$. Horizontal lines show the ``true'' number of clusters.}
    \label{fig:nc_alpha_sens}
\end{figure}

In our experiments we select $\alpha$ proportional to the estimated standard deviation of the random variable $\v^\top X$, where $\v$ parameterises the minimum density hyperplane $H(\v, b)$. Figures~\ref{fig:nmi_alpha_sens} and~\ref{fig:nc_alpha_sens} show plots of the NMI and number of clusters selected for varying $\alpha$ between 0 and 2 times this standard deviation. While the number of clusters selected is similar for different values of $\alpha$, the performance appears most consistently good for smaller values, i.e., reasonably balanced splits of the data. However, this is in the context of assessment based on NMI, where different external validation metrics penalise over/under-splitting of clusters to different extents. In the context of large data sets, it seems reasonable that over-splitting is relatively acceptable since reasonable inferences can be made about even subsets of ``true clusters'', as these will tend to be large enough in sample size.

%\subsection{Dimensionality Reduction}

\section{Conclusions}

Low density hyperplanes are intuitively appealing as cluster separators and have been successfully applied in numerous areas, including high dimensional applications. In this work we presented a fully incremental approach for obtaining low density hyperplanes based on stochastic gradient descent applied to a convolution of a sequence of i.i.d. random variables with a smoothing kernel with decreasing bandwidth. We showed that such an approach leads to convergence to a stationary point in the minimum density hyperplane objective under very mild assumptions which are all satisfied by finite Gaussian mixtures.
Using a simple offline pruning of a hierarchical binary tree model formed by low density hyperplane separators we found that this approach leads to efficient and accurate clustering in various applications, and shows favourable performance to existing incremental/online clustering methods and is competitive with batch implementations of $k$-means and dePDDP.

\newpage

\section*{Appendix}

\subsection*{Assumption 1 holds for finite Gaussian mixtures}

Suppose that $X$ has a $K$ component Gaussian mixture distribution on $\R^d$, with mixing proportions $\pi_1, ..., \pi_K$ and component means and covariances respectively given by $\mmu_1, ..., \mmu_K$ and $\Sigma_1, ..., \Sigma_K$. Then for any $\v \in \R^d$ the random variable $\v^\top X$ has a $K$ component Gaussian mixture distribution on $\R$ with the same mixing proportions, and with component means and variances given respectively by $\v^\top\mmu_1, ..., \v^\top\mmu_K$ and $\v^\top\Sigma_1\v, ..., \v^\top\Sigma_K\v$. We can therefore write
\begin{align*}
    f_{\v^\top X}(b) &= \sum_{k=1}^K \pi_k f^{(k)}_{\v^\top X}(b),\\
    f^{(k)}_{\v^\top X}(b) &:= \frac{1}{\sqrt{2\pi \v^\top \Sigma_k\v}}\exp\left(-\frac{(b - \v^\top\mmu_k)^2}{2\v^\top\Sigma_k\v}\right).
\end{align*}
We therefore find that
\begin{align*}
    \nabla_\v f^{(k)}_{\v^\top X}(b) &= \frac{1}{\sqrt{2\pi}(\v^\top \Sigma_k \v)^{3/2}}\exp\left(-\frac{(b - \v^\top\mmu_k)^2}{2\v^\top\Sigma_k \v}\right)\left((b - \v^\top\mmu_k)\mmu_k + \left(\frac{(b-\v^\top\mmu_k)^2}{\v^\top\Sigma_k\v} - 1\right)\Sigma_k\v\right)\\
    \frac{\partial}{\partial b}f^{(k)}_{\v^\top X}(b) &= \frac{1}{\sqrt{2\pi}(\v^\top \Sigma_k \v)^{3/2}}\exp\left(-\frac{(b - \v^\top\mmu_k)^2}{2\v^\top\Sigma_k \v}\right)(\v^\top\mmu_k - b).
\end{align*}
Now take any $\v_1, \v_2 \in \R^{d}$ and $b_1, b_2 \in \R$ and set
\begin{align*}
    \v_1^\top\Sigma_k\v_1 - \v_2^\top\Sigma_k\v_2 &=: \epsilon_1,\\
    \sqrt{\v_1^\top\Sigma_k\v_1} - \sqrt{\v_2^\top\Sigma_k\v_2} &=: \epsilon_2,\\
    (b_1-\v_1^\top\mmu_k) - (b_2-\v_2^\top\mmu_k) &=: \epsilon_3.
\end{align*}
From these we get
\begin{align*}
    2\v_2^\top\Sigma_k(\v_1-\v_2) + (\v_1-\v_2)^\top\Sigma_k(\v_1-\v_2) = \epsilon_1 = 2\sqrt{\v_1^\top\Sigma_k\v_1}\epsilon_2 - \epsilon_2^2.
\end{align*}
Now, it can be verified that $\exp(-x^2) \leq \exp(-y^2) -2y\exp(-y^2)(x-y) + \frac{16}{1+y^2}(x-y)^2$ for all $x, y \in \R$. We therefore have
%, which arises from a second order Taylor expansion and the fact that $\frac{d^2}{d x^2}\exp(-x^2) \leq 1$ for all $x$, we get
%
\begin{align*}
    \frac{\exp\left(-\frac{(b_2-\v_2^\top\mmu_k)^2}{2\v_2^\top\Sigma_k\v_2}\right)}{\sqrt{2\pi \v_2^\top\Sigma_k\v_2}} \leq&  \frac{\exp\left(-\frac{(b_1-\v_1^\top\mmu_k)^2}{2\v_1^\top\Sigma_k\v_1}\right)}{\sqrt{2\pi \v_1^\top\Sigma_k\v_1}}\Bigg(\sqrt{\frac{\v_1^\top\Sigma_k\v_1}{\v_2^\top\Sigma_k\v_2}} - \frac{b_1 - \v_1^\top\mmu_k}{\sqrt{\v_2^\top\Sigma_k\v_2}}\left(\frac{b_2 - \v_2^\top\mmu_k}{\sqrt{\v_2^\top\Sigma_k\v_2}} - \frac{b_1 - \v_1^\top\mmu_k}{\sqrt{\v_1^\top\Sigma_k\v_1}}\right)\Bigg)\\
    & + \frac{16}{\sqrt{2\pi\v_2^\top\Sigma_k\v_2}\left(1+\frac{(b_1-\v_1^\top\mmu_k)^2}{2\v_1^\top\Sigma_k\v_1}\right)}\left(\frac{b_1 - \v_1^\top\mmu_k}{\sqrt{2\v_1^\top\Sigma_k\v_1}} - \frac{b_2 - \v_2^\top\mmu_k}{\sqrt{2\v_2^\top\Sigma_k\v_2}}\right)^2.
\end{align*}
Now, since $\sqrt{x}$ is concave and its dervative at 1 is 1/2 we have $\sqrt{\v_1^\top\Sigma_k\v_1/\v_2^\top\Sigma\v_2} \leq 1 + \epsilon_1/2\v_2^\top\Sigma_k\v_2$. Furthermore, we have
\begin{align*}
    \frac{b_1 - \v_1^\top\mmu_k}{\sqrt{\v_2^\top\Sigma_k\v_2}}&\left(\frac{b_2 - \v_2^\top\mmu_k}{\sqrt{\v_2^\top\Sigma_k\v_2}} - \frac{b_1 - \v_1^\top\mmu_k}{\sqrt{\v_1^\top\Sigma_k\v_1}}\right) = - \frac{b_1 - \v_1^\top\mmu_k}{\v_2^\top\Sigma_k\v_2}\epsilon_3 + \frac{(b_1-\v_1^\top\mmu_k)^2}{\v_2^\top\Sigma_k\v_2\sqrt{\v_1^\top\Sigma_k\v_1}}\epsilon_2\\
    &= -\frac{b_1 - \v_1^\top\mmu_k}{\v_1^\top\Sigma_k\v_1}\epsilon_3 - \frac{b_1 - \v_1^\top\mmu_k}{(\v_1^\top\Sigma_k\v_1)(\v_2^\top\Sigma_k\v_2)}\epsilon_1\epsilon_3 + \frac{(b_1-\v_1^\top\mmu_k)^2}{2(\v_2^\top\Sigma_k\v_2)(\v_1^\top\Sigma_k\v_1)}(\epsilon_1+ \epsilon_2^2)
\end{align*}
In all we therefore find that
\begin{align*}
    f^{(k)}_{\v_2^\top X}(b_2) \leq & f^{(k)}_{\v_1^\top X}(b_1) +  \frac{\exp\left(-\frac{(b_1-\v_1^\top\mmu_k)^2}{2\v_1^\top\Sigma_k\v_1}\right)}{\sqrt{2\pi \v_1^\top\Sigma_k\v_1}}\Bigg(\frac{\epsilon_1}{2\v_2^\top\Sigma_k\v_2} + \frac{b_1 - \v_1^\top\mmu_k}{\v_1^\top\Sigma_k\v_1}\epsilon_3\\
    & + \frac{b_1 - \v_1^\top\mmu_k}{(\v_1^\top\Sigma_k\v_1)(\v_2^\top\Sigma_k\v_2)}\epsilon_1\epsilon_3 -\frac{(b_1-\v_1^\top\mmu_k)^2}{2(\v_2^\top\Sigma_k\v_2)(\v_1^\top\Sigma_k\v_1)}(\epsilon_1+ \epsilon_2^2) \Bigg)\\
    & + \frac{16}{\sqrt{2\pi\v_2^\top\Sigma_k\v_2}\left(1+\frac{(b_1-\v_1^\top\mmu_k)^2}{2\v_1^\top\Sigma_k\v_1}\right)}\left(\frac{b_1 - \v_1^\top\mmu_k}{\sqrt{2\v_1^\top\Sigma_k\v_1}} - \frac{b_2 - \v_2^\top\mmu_k}{\sqrt{2\v_2^\top\Sigma_k\v_2}}\right)^2.
\end{align*}
Now consider that
\begin{align*}
    \nabla_{(\v, b)} f^{(k)}_{\v_1^\top X}(b_1)^\top&\left((\v_2, b_2) - (\v_1, b_1)\right) =\\
    &\frac{\exp\left(-\frac{(b_1-\v_1^\top\mmu_k)^2}{2\v_1^\top\Sigma_k\v_1}\right)}{\sqrt{2\pi \v_1^\top\Sigma_k\v_1}}\Bigg(\frac{b_1-\v_1^\top\mmu_k}{\v_1^\top\Sigma_k\v_1}\epsilon_3  + \left(\frac{(b_1-\v_1^\top\mmu_k)^2}{\v_1^\top\Sigma_k\v_1} - 1\right)\frac{\v_1^\top\Sigma_k(\v_2-\v_1)}{\v_1^\top\Sigma_k\v_1}\Bigg).
\end{align*}
Finally, observe that
\begin{align*}
    \frac{\v_1^\top\Sigma_k(\v_2-\v_1)}{\v_1^\top\Sigma_k\v_1} &= -\frac{\epsilon_1 - (\v_1-\v_2)^\top\Sigma_k(\v_1-\v_2)}{2\v_1^\top\Sigma_k\v_1}\\
    &= -\frac{\epsilon_1}{2\v_2^\top\Sigma_k\v_2} + \frac{\epsilon_1^2}{2(\v_1^\top\Sigma_k\v_1)(\v_2^\top\Sigma_k\v_2)} + \frac{(\v_1-\v_2)^\top\Sigma_k(\v_1-\v_2)}{2\v_1^\top\Sigma_k\v_1}.
\end{align*}
Combining all of this we have
\begin{align*}
    f^{(k)}_{\v_2^\top X}(b_2)& - f^{(k)}_{\v_1^\top X}(b_1) - \nabla_{(\v, b)} f^{(k)}_{\v_1^\top X}(b_1)^\top\left((\v_2, b_2) - (\v_1, b_1)\right) \leq\\
    &\frac{\exp\left(-\frac{(b_1-\v_1^\top\mmu_k)^2}{2\v_1^\top\Sigma_k\v_1}\right)}{\sqrt{2\pi \v_1^\top\Sigma_k\v_1}}\Bigg(\frac{b_1-\v_1^\top\mmu_k}{(\v_1^\top\Sigma_k\v_1)(\v_2^\top\Sigma_k\v_2)}\epsilon_1\epsilon_3 - \frac{(b_1-\v_1^\top\mmu_k)^2}{2(\v_2^\top\Sigma_k\v_2)(\v_1^\top\Sigma_k\v_1)}\epsilon_2^2\Bigg)\\
    &+ \frac{16}{\sqrt{2\pi\v_2^\top\Sigma_k\v_2}\left(1+\frac{(b_1-\v_1^\top\mmu_k)^2}{2\v_1^\top\Sigma_k\v_1}\right)}\left(\frac{b_1 - \v_1^\top\mmu_k}{\sqrt{2\v_1^\top\Sigma_k\v_1}} - \frac{b_2 - \v_2^\top\mmu_k}{\sqrt{2\v_2^\top\Sigma_k\v_2}}\right)^2\\
    \leq& \frac{\exp\left(-\frac{(b_1-\v_1^\top\mmu_k)^2}{2\v_1^\top\Sigma_k\v_1}\right)}{\sqrt{2\pi \v_1^\top\Sigma_k\v_1}}\frac{b_1-\v_1^\top\mmu_k}{(\v_1^\top\Sigma_k\v_1)(\v_2^\top\Sigma_k\v_2)}\epsilon_1\epsilon_3\\
    &+ \frac{16}{\sqrt{2\pi\v_2^\top\Sigma_k\v_2}\left(1+\frac{(b_1-\v_1^\top\mmu_k)^2}{2\v_1^\top\Sigma_k\v_1}\right)}\left(\frac{b_1 - \v_1^\top\mmu_k}{\sqrt{2\v_1^\top\Sigma_k\v_1}} - \frac{b_2 - \v_2^\top\mmu_k}{\sqrt{2\v_2^\top\Sigma_k\v_2}}\right)^2\\
    =& \frac{\exp\left(-\frac{(b_1-\v_1^\top\mmu_k)^2}{2\v_1^\top\Sigma_k\v_1}\right)}{\sqrt{2\pi \v_1^\top\Sigma_k\v_1}}\frac{b_1-\v_1^\top\mmu_k}{(\v_1^\top\Sigma_k\v_1)(\v_2^\top\Sigma_k\v_2)}\epsilon_1\epsilon_3 + \frac{16\left(1+\frac{(b_1-\v_1^\top\mmu_k)^2}{2\v_1^\top\Sigma_k\v_1}\right)^{-1}}{\sqrt{\pi}(2\v_2^\top\Sigma_k\v_2)^{3/2}}\\
    &\times\Bigg((b_1-b_2 + (\v_2-\v_1)^\top\mmu_k)^2+ (b_1 - \v_1^\top\mmu_k)^2\left(\sqrt{\frac{\v_2^\top\Sigma_k\v_2}{\v_1^\top\Sigma_k\v_1}} - 1\right)^2\\
    & \hspace{40pt}+ 2(b_1 - \v_1^\top\mmu_k)\left(\sqrt{\frac{\v_2^\top\Sigma_k\v_2}{\v_1^\top\Sigma_k\v_1}} - 1\right)(b_1-b_2+(\v_2-\v_1)^\top\mmu_k)\Bigg)\\
    \leq & \frac{\exp\left(-\frac{(b_1-\v_1^\top\mmu_k)^2}{2\v_1^\top\Sigma_k\v_1}\right)}{\sqrt{2\pi \v_1^\top\Sigma_k\v_1}}\frac{b_1-\v_1^\top\mmu_k}{(\v_1^\top\Sigma_k\v_1)(\v_2^\top\Sigma_k\v_2)}\epsilon_1\epsilon_3 + \frac{16\left(1+\frac{(b_1-\v_1^\top\mmu_k)^2}{2\v_1^\top\Sigma_k\v_1}\right)^{-1}}{\sqrt{\pi}(2\v_2^\top\Sigma_k\v_2)^{3/2}}\\
    &\times\bigg((b_1-b_2)^2 + ||\v_1-\v_2||^2||\mmu_k||^2 + |b_1-b_2|\cdot||\v_1-\v_2|| + \frac{(b_1-\v_1^\top\mmu_k)^2}{4(\v_1^\top\Sigma_k\v_1)^2}\epsilon_1^2\\
    &\hspace{140pt} + 2\frac{|b_1-\v_1^\top\mmu_k|}{2\v_1^\top\Sigma_k\v_1}|\epsilon_1|(|b_1-b_2| + ||\v_1-\v_2||\cdot||\mmu_k||)\bigg).
\end{align*}
Now, since $\exp(-(b - \v^\top\mmu_k)^2/2\v^\top\Sigma_k\v)(b - \v^\top\mmu_k)$ is bounded above uniformly in $\v, b$ and $\v^\top\Sigma_k\v$ is bounded above and below for $||\v|| = 1$ since we assume $\Sigma_k$ has full rank, we find that the above is bounded by
\begin{align*}
    D^{(k)}\left(||\v_1-\v_2||^2 + (b_1-b_2)^2+|b_1-b_2|\cdot ||\v_1-\v_2||\right),
\end{align*}
for some constant $D^{(k)}$ independent of $\v_1, \v_2, b_1, b_2$, assuming $||\v_1||, ||\v_2|| = 1$.

Since $f_{\v^\top X}(b) = \sum_{k=1}^K \pi_k f^{(k)}_{\v^\top X}(b)$, where $\sum_{k=1}^K \pi_k = 1$, we therefore have
\begin{align*}
    f_{\v_2^\top X}(b_2)& - f_{\v_1^\top X}(b_1) - \nabla_{(\v, b)} f_{\v_1^\top X}(b_1)^\top\left((\v_2, b_2) - (\v_1, b_1)\right) \leq\\
    &\max_{k} \{D^{(k)}\}\left(||\v_1-\v_2||^2 + (b_1-b_2)^2+|b_1-b_2|\cdot ||\v_1-\v_2||\right),
\end{align*}
as required.

\subsection*{Bounding $E[\epsilon_\v^{(t+1)}|\X^{(1:t)}]$ in Eq.~(\ref{eq:eps_error})}

Here we investigate the magnitude of the bias of the stochastic gradients of the objective with respect to the sequence of vectors $\v^{(0)}, \v^{(1)}, ...$. In order to do so it is first important to formalise how exactly we can express the partial derivatives of the objective $f_{\v^\top X}(b) = I_{f}(\v, b)$ with respect to the elements in $\v$. We begin by re-expressing the surface integral as
\begin{align*}
    f_{\v^\top X}(b) = \oint_{\x:\v^\top \x = b} f(\x) d\x = \lim_{\epsilon \to 0^+} \frac{1}{2\epsilon} P\left(b - \epsilon \leq \v^\top X \leq b + \epsilon\right).
\end{align*}
Now, if we let $\v_{-i}$ and $X_{-i}$ be $\v$ and $X$ but excluding the $i$-th entry, we have
\begin{align*}
    P\left(b - \epsilon \leq \v^\top X \leq b + \epsilon\right) &= P\left(b - \epsilon - v_iX_i \leq \v_{-i}^\top X_{-i} \leq b + \epsilon - v_iX_i\right)\\
    &= \int_{-\infty}^\infty P\left(b - \epsilon - v_ix \leq \v_{-i}^\top X_{-i} \leq b + \epsilon - v_ix | X_i = x\right)f_{X_i}(x) dx\\
    \Rightarrow \lim_{\epsilon \to 0^+}\frac{P\left(b - \epsilon \leq \v^\top X \leq b + \epsilon\right)}{2\epsilon} &= \int_{-\infty}^\infty f_{\v_{-i}^\top X_{-i}}(b - v_ix|X_i = x)f_{X_i}(x) dx.
\end{align*}
Exactly analogously, if $\e_i$ is the $i$-th canonical basis vector for $\R^d$, then
\begin{align*}
    \lim_{\epsilon \to 0^+}\frac{P\left(b - \epsilon \leq (\v+h\e_i)^\top X \leq b + \epsilon\right)}{2\epsilon} &= \int_{-\infty}^\infty f_{\v_{-i}^\top X_{-i}}(b - v_ix - hx|X_i = x)f_{X_i}(x) dx.
\end{align*}
We therefore have
\begin{align*}
    \frac{\partial f_{\v}(b)}{\partial v_i} &= \lim_{h\to 0}\frac{1}{h} \int_{-\infty}^\infty \left(f_{\v_{-i}^\top X_{-i}}(b - v_ix - hx|X_i = x) - f_{\v_{-i}^\top X_{-i}}(b - v_ix|X_i = x)\right)f_{X_i}(x) dx\\
    &= \int_{-\infty}^\infty \lim_{h\to 0}\frac{1}{h} \left(f_{\v_{-i}^\top X_{-i}}(b - v_ix - hx|X_i = x) - f_{\v_{-i}^\top X_{-i}}(b - v_ix|X_i = x)\right)f_{X_i}(x) dx\\
    &= -\int_{-\infty}^\infty  xf_{\v_{-i}^\top X_{-i}}^\prime(b - v_ix|X_i = x)f_{X_i}(x) dx.
\end{align*}
Now, consider the $i$-th element of the stochastic gradient, i.e., $\frac{b - \v^\top X}{h^3}\phi\left(\frac{b - \v^\top X}{h}\right)X_i$. We have,
\begin{align*}
    E\left[\frac{b - \v^\top X}{h^3}\phi\left(\frac{b - \v^\top X}{h}\right)X_i\right] &= E_{X_i}\left[E\left[\frac{b - \v^\top X}{h^3}\phi\left(\frac{b - \v^\top X}{h}\right)X_i\bigg| X_i \right]\right],\\
    E\left[\frac{b - \v^\top X}{h^3}\phi\left(\frac{b - \v^\top X}{h}\right)X_i\bigg| X_i = x \right]&= x \int_{-\infty}^\infty \frac{b - v_i x - z}{h^3}\phi\left(\frac{b - v_i x - z}{h}\right) f_{\v_{-i}^\top X_{-i}}(z|X_i = x) dz.
\end{align*}
Now, using a standard transformation of variables as is common in investigating the error of kernel smoothing methods, let $u = (b - v_ix - z)/h$ and hence $z = b - v_i x - hu$, $|dz| = h|du|$. The integral above is therefore equal to
\begin{align*}
    \int_{-\infty}^\infty & \frac{u}{h}\phi(u)f_{\v_{-i}^\top X_{-i}}(b - v_i x - hu|X_i = x) du\\
    &= \int_{-\infty}^\infty \frac{u}{h}\phi(u)\bigg(f_{\v_{-i}^\top X_{-i}}(b - v_i x|X_i = x) - huf_{\v_{-i}^\top X_{-i}}^\prime (b - v_i x|X_i = x)\\
    & \hspace{20pt} + \frac{h^2u^2}{2}f_{\v_{-i}^\top X_{-i}}^{\prime\prime} (b - v_i x|X_i = x) - \frac{h^3u^3}{6}f_{\v_{-i}^\top X_{-i}}^{\prime\prime\prime} (b - v_i x|X_i = x) + o(h^3u^3) \bigg) du\\
    &= -f_{\v_{-i}^\top X_{-i}}^\prime (b - v_i x|X_i = x) + \mathcal{O}(h^2),
\end{align*}
since $\int_{-\infty}^\infty u^k\phi(u) du = 0$ for all odd $k$ and $\int_{-\infty}^\infty u^2\phi(u)du = 1$. %
%Note also that the magnitude of the constant suppressed in $\mathcal{O}(h^2)$ may be chosen independent of $(\v, b)$ by assumption 1 in Section~\ref{sec:convergence}.
%
In all we therefore have,
\begin{align*}
    E\left[\frac{b - \v^\top X}{h^3}\phi\left(\frac{b - \v^\top X}{h}\right)X_i\right] &= - E\left[X_i\left(f_{\v_{-i}^\top X_{-i}}^\prime (b - v_i X_i|X_i)+ \mathcal{O}(h^2)\right)\right]\\
    &= -\int_{-\infty}^\infty  xf_{\v_{-i}^\top X_{-i}}^\prime(b - v_ix|X_i = x)f_{X_i}(x) dx + \mathcal{O}(h^2)\\
    &= \frac{\partial}{\partial v_i} f_{\v}(b) + \mathcal{O}(h^2).
\end{align*}
In the above we have treated $\v, b$ and $h$ as fixed, however since $\v^{(t)}, b^{(t)}$ and $h^{(t+1)}$ are fully determined by $\X^{(1:t)}$, we have
\begin{align*}
    E\left[\frac{b^{(t)} - \v^{(t)\top} X^{(t+1)}}{(h^{(t+1)})^3}\phi\left(\frac{b^{(t)} - \v^{(t+1)\top} X^{(t+1)}}{h^{(t+1)}}\right)X^{(t+1)} \big| \X^{(1:t)}\right] - \nabla_{\v} f_{\v^{(t)\top} X}(b^{(t)})&= \mathcal{O}(t^{-2q}),
\end{align*}
since $h^{(t+1)} = s^{(t)}(t+1)^{-q}$ with $s^{(t)}$ almost surely bounded above and away from zero.

\subsection*{Bounding $E\left[\frac{\partial}{\partial b}O(\v^{(t)}, b^{(t)} )\epsilon_b^{(t+1)}\right]$ in Eq.~(\ref{eq:dbeps_error})}

First recall that 
\begin{align*}
    \frac{\partial}{\partial b} O(\v, b) &= \frac{\partial}{\partial b} f_{\v^\top X}(b) + 2C(|b|-\alpha)_+\mathrm{sign}(b)\\
    &= f_{\v^\top X}^\prime(b) + 2C(|b|-\alpha)_+\mathrm{sign}(b).
\end{align*}
Now, using an almost exactly analogous approach to that in previous derivation, we have
\begin{align*}
    E\bigg[&\frac{b - \v^\top X}{h^3}\phi\left(\frac{b - \v^\top X}{h}\right)\bigg] = \int_{-\infty}^\infty \frac{u}{h}\phi(u) f_{\v^\top X}(b+hu) du\\
    &= \int_{-\infty}^\infty \frac{u}{h}\phi(u) \left(f_{\v^\top X}(b) + huf^\prime_{\v^\top X}(b) + \frac{(uh)^2}{2}f^{\prime\prime}_{\v^\top X}(b) + \frac{(uh)^3}{6}f^{\prime\prime\prime}_{\v^\top X}(b)   + o(h^3u^3)\right) du\\
    &= f^\prime_{\v^\top X}(b) + \frac{E[N^4]}{6}h^2f^{\prime\prime\prime}_{\v^\top X}(b) + o(h^3),
\end{align*}
where $E[N^4]$ is the fourth moment of a univariate standard Gaussian random variable.
Then consider that for $t > M^{1/\eta}$, with $M$ given in Assumption 4 in Section~\ref{sec:convergence}, we have
\begin{align*}
    E&\bigg[(|b^{(t)}|-\alpha)_+\sgn(b^{(t)})\left(\frac{b^{(t)}-\v^{(t)\top} X^{(t+1)} }{(h^{(t+1)})^3}\phi\left(\frac{b^{(t)} - \v^{(t)\top} X^{(t+1)}}{h^{(t+1)}}\right)-f^\prime_{\v^{(t)^\top} X}(b^{(t)})\right)\big| \v^{(t)}, h^{(t+1)}\bigg]\\
    &= E\Bigg[(|b^{(t)}|-\alpha)_+\sgn(b^{(t)})E\bigg[\frac{b^{(t)}-\v^{(t)\top} X^{(t+1)} }{(h^{(t+1)})^3}\phi\left(\frac{b^{(t)} - \v^{(t)\top} X^{(t+1)}}{h^{(t+1)}}\right)\\
    &\hspace{160pt}-f^\prime_{\v^{(t)^\top} X}(b^{(t)})\big| \v^{(t)}, h^{(t+1)}, b^{(t)}\bigg] \bigg | \v^{(t)}, h^{(t+1)}\Bigg]\\
    & = E\left[(|b^{(t)}|-\alpha)_+\sgn(b^{(t)})\bigg(\frac{E[N^4]}{6}(h^{(t+1)})^2f^{\prime\prime\prime}_{\v^{(t)\top} X}(b^{(t)}) + o((h^{(t+1)})^3)\bigg) \bigg | \v^{(t)}, h^{(t+1)}\right]\\
    &= \int_{-\infty}^\infty (|z|-\alpha)_+\sgn(z)\bigg(\frac{E[N^4]}{6}(h^{(t+1)})^2f^{\prime\prime\prime}_{\v^{(t)\top} X}(z) + o((h^{(t+1)})^3)\bigg) f_{b^{(t)}}(z|\v^{(t)}, h^{(t+1)}) dz\\
    &\leq \frac{E[N^4]}{6}(h^{(t+1)})^2 \int_{-\infty}^{\infty}|zf^{\prime\prime\prime}_{\v^{(t)\top} X}(z)| f_{b^{(t)}}(z|\v^{(t)}, h^{(t+1)}) dz + o((h^{(t+1)})^3)\\
    &= \frac{E[N^4]}{6}(h^{(t+1)})^2\Bigg( \int_{|z| \leq t^{\eta}}|zf^{\prime\prime\prime}_{\v^{(t)\top} X}(z)| f_{b^{(t)}}(z|\v^{(t)}, h^{(t+1)}) dz\\
    &\hspace{100pt} + \int_{|z| > t^{\eta}}|zf^{\prime\prime\prime}_{\v^{(t)\top} X}(z)| f_{b^{(t)}}(z|\v^{(t)}, h^{(t+1)}) dz\Bigg) + o((h^{(t+1)})^3)\\
    &\leq D_1 t^\eta (h^{(t+1)})^2 + D_2 t^{-\eta} + D_3 (h^{(t+1)})^3,
\end{align*}
for some constants $D_1, D_2, D_3$ independent of $\v^{(t)}, h^{(t+1)}$, using Assumptions 3 and 4 from Section~\ref{sec:convergence}.

Since $h^{(t+1)} = s^{(t)}(t+1)^{-q}$, we therefore have
\begin{align*}
    E&\left[\frac{\partial}{\partial b} O(\v^{(t)}, b^{(t)}) \epsilon_b^{(t+1)}\right] \\
    &= E\left[\left(f^\prime_{\v^{(t)^\top} X}(b^{(t)}) + 2C(|b^{(t)}|-\alpha)_+\sgn(b^{(t)})\right)\left(\frac{b^{(t)}-\v^{(t)\top} X^{(t+1)} }{(h^{(t+1)})^3}\phi\left(\frac{b^{(t)} - \v^{(t)\top} X^{(t+1)}}{h^{(t+1)}}\right)-f^\prime_{\v^\top X}(b)\right)\right]\\
    &\leq E[D_4(h^{(t+1)})^2 + D_1 t^\eta (h^{(t+1)})^2 + D_2 t^{-\eta} + D_3 (h^{(t+1)})^3] = \mathcal{O}(t^{\eta - 2q} + t^{-\eta}),
\end{align*}
where $D_4$ is some constant coming from the fact that $f^{\prime}_{\v^\top X}(b)$ is bounded w.r.t. $(\v, b)$.

\subsection*{Bounding $E[||\u^{(t+1)}||^2|\X^{(1:t)}]$ in Eq.~(\ref{eq:u^2_error}).}

This bound arises trivially from the fact that $u\phi(u)$ is bounded, and hence
\begin{align*}
    ||\u^{(t+1)}||^2 &= \left\|\frac{b^{(t)}-\v^{(t)\top}X^{(t+1)}}{(h^{(t+1)})^3}\phi\left(\frac{b^{(t)}-\v^{(t)\top}X^{(t+1)}}{h^{(t+1)}}\right)X^{(t+1)}\right\|^2\\
    &\leq D(h^{(t+1)})^{-4}||X^{(t+1)}||^2,
\end{align*}
for some constant $D$ indepentent of $t$, and so $E[||X||^2] < \infty$ we have
\begin{align*}
    E[||\u^{(t+1)}||^2|\X^{(1:t)}] &= \mathcal{O}((t+1)^{4q}) = \mathcal{O}(t^{4q}).
\end{align*}

\subsection*{Bounding $E[(b^{(t+1)}-b^{(t)})^2 | \X^{(1:t)}]$ in Eq.~(\ref{eq:bmb_error}).}

We first show by a quick induction that $|b^{(t)}| \leq |b^{(0)}|a^t + c \frac{1-a^t}{1-a}$, where $a = \max\{1, 2\bar\gamma_2C\}$ and $c = \bar\gamma_2 K/Q$ where $K$ is an upper bound for $|u\phi(u)|$ and $Q>0$ is an almost sure lower bound for the sequence $s^{(0)}, s^{(1)}, ...$. In this we take $0/0$ to be equal to 0. This inequality holds trivially for $t=0$ as all constants are positive. Suppose then that it holds for all $0\leq t \leq k$ for some $k$. Consider
\begin{align*}
    b^{(k+1)} &= b^{(k)} - \bar\gamma_2(k+1)^{-r}\left(\frac{b^{(k)}-\v^{(k)^\top} X^{(k+1)}}{(h^{(k+1)})^3}\phi\left(\frac{b^{(k)}-\v^{(k)^\top} X^{(k+1)}}{h^{(k+1)}}\right) + 2C(|b^{(k)}|-\alpha)_+\sgn(b^{(k)})\right)\\
    &= b^{(k)}\left(1-2C\bar\gamma_2\frac{(|b^{(k)}|-\alpha)_+}{|b^{(k)}|}(k+1)^{-r}\right) - \bar\gamma_2(k+1)^{-r}\frac{b^{(k)}-\v^{(k)^\top} X^{(k+1)}}{(h^{(k+1)})^3}\phi\left(\frac{b^{(k)}-\v^{(k)^\top} X^{(k+1)}}{h^{(k+1)}}\right).
\end{align*}
Clearly $(|b^{(k)}|-\alpha)_+/|b^{(k)}| \leq 1$, and since $h^{(k+1)} \geq Q(k+1)^{-q}$ and $r > 2q$, we can therefore write
\begin{align*}
    |b^{(k+1)}| &\leq a|b^{(k)}| + c\\
    &\leq a\left(|b^{(0)}|a^k + c\frac{1-a^k}{1-a}\right) + c\\
    &= |b^{(0)}|a^{k+1} + c\left(a\frac{1-a^k}{1-a} + 1\right)\\
    &= |b^{(0)}|a^{k+1} + c\frac{1-a^{k+1}}{1-a},
\end{align*}
and hence the induction holds. 

Now, since $|b^{(0)}|a^t + c \frac{1-a^t}{1-a}$ is non-decreasing in $t$ we can conclude that for all $t \leq t^* := (\bar\gamma_2C)^{1/r}$ we have $|b^{(t)}|\leq B:= |b^{(0)}|a^{t^*} + c\frac{1-a^{t^*}}{1-a}$. Now $t^*$ is chosen so that for $t \geq t^*$ we have $ \gamma_2^{(t+1)}C(|b^{(t)}|-\alpha)_+ \leq |b^{(t)}|$, and hence
\begin{align*}
    |b^{(t+1)}| &= \left|b^{(t)} - \bar\gamma_2(t+1)^{-r}\left(\frac{b^{(t)}-\v^{(t)\top} X^{(t+1)}}{(h^{(t+1)})^3}\phi\left(\frac{b^{(t)}-\v^{(t)^\top} X^{(t+1)}}{h^{(t+1)}}\right) + 2C(|b^{(t)}|-\alpha)_+\sgn(b^{(t)})\right)\right|\\
    & \leq |b^{(t)}| + c (t+1)^{-(r-2q)} \leq B + c\sum_{j=t^*}^{t} (j+1)^{-(r-2q)}\\
    & \leq B + c\sum_{j=0}^{t} (j+1)^{-(r-2q)} \leq B + \frac{c}{1-(r-2q)}(t+1)^{1-(r-2q)}.
\end{align*}
The second term is clearly positive since $0 < r-2q < 1$ and hence we know that for all $t$ we have $|b^{(t)}| \leq B + \frac{c}{1-(r-2q)}t^{1-(r-2q)},$ and hence $|b^{(t)}|$ cannot grow faster than $\mathcal{O}(t^{1-(r-2q)})$ for $t\geq t^*$. Combining all of this we have for $t\geq t^*$ that
\begin{align*}
    |b^{(t+1)} - b^{(t)}| &= \bar \gamma_2(t+1)^{-r}\left|\frac{b^{(t)}-\v^{(t)\top} X^{(t+1)}}{(h^{(t+1)})^3}\phi\left(\frac{b^{(t)}-\v^{(t)^\top} X^{(t+1)}}{h^{(t+1)}}\right) + 2C(|b^{(t)}|-\alpha)_+\sgn(b^{(t)})\right|\\
    & \leq c(t+1)^{-(r-2q)} + (t+1)^{-r}2C\left(B + \frac{c}{1-(r-2q)}(t+1)^{1-(r-2q)}\right)\\
    &= \mathcal{O}(t^{-(r-2q)}),
\end{align*}
since $r \leq 1$. Trivially then we have $E[(b^{(t+1)}-b^{(t)})^2 | \X^{(1:t)}] = \mathcal{O}(t^{-2(r-2q)})$.

\subsection*{Bounding $E[||\v^{(t+1)}-\v^{(t)}||^2 | \X^{(1:t)}]$ in Eq.~(\ref{eq:vmv_error}).}

Recall first that
\begin{align*}
    \v^{(t+1)} - \v^{(t)} &= - \bar \gamma_1 (t+1)^{-r} (\I - \v^{(t)}\v^{(t)\top}) \u^{(t+1)} - \mathcal{O}((t+1)^{-2r}(\v^{(t)\top}\u^{(t+1)})^2)
\end{align*}
and hence
\begin{align*}
    E[||\v^{(t+1)} - \v^{(t)}||^2 | \X^{(1:t)}] &\leq  \bar \gamma_1^2 (t+1)^{-2r}E[||\u^{(t+1)}||^2| \X^{(1:t)}] + o(E[||\u^{(t+1)}||^2]t^{-2r})\\
    &= \mathcal{O}(t^{-2r}E[||\u^{(t+1)}||^2| \X^{(1:t)}])\\
    &= \mathcal{O}(t^{-2(r-2q)}),
\end{align*}
since the largest eigenvalue of $\I-\v^{(t)}\v^{(t)\top}$ is one, and hence $||(\I-\v^{(t)}\v^{(t)\top})\u^{(t+1)}||^2 \leq ||\u^{(t+1)}||^2$.

\subsection*{Bounding $E\left[|b^{(t+1)} - b^{(t)}|\cdot ||\v^{(t+1)}-\v^{(t)}|| \ | \X^{(1:t)}\right]$ in Eq.~(\ref{eq:bmbvmv_error}).}

It follows immediately from the fact that $|b^{(t+1)} - b^{(t)}| = \mathcal{O}(t^{-(r-2q)})$ deterministically and that $E[||\v^{(t+1)} - \v^{(t)}||^2 | \X^{1:t}] = \mathcal{O}(t^{-2(r-2q)})$, plus the fact that $E[Z] \leq \sqrt{E[Z^2]}$ for any random variable $Z$, that we have $E\left[|b^{(t+1)} - b^{(t)}|\cdot ||\v^{(t+1)}-\v^{(t)}|| \ | \X^{(1:t)}\right] = \mathcal{O}(t^{-2(r-2q)})$.

\bibliographystyle{plainnat}

\bibliography{Bibliography-MM-MC}

%% -- Appendix (if any) --------------------------------------------------------
%% - After the bibliography with page break.
%% - With proper section titles and _not_ just "Appendix".

\end{document}